\documentclass{article}

\PassOptionsToPackage{numbers,square,sort&compress}{natbib}
\usepackage[preprint]{neurips_2026}

\usepackage[utf8]{inputenc}
\usepackage[T1]{fontenc}
\usepackage{url}
\usepackage{booktabs}
\usepackage{amsfonts}
\usepackage{amssymb}
\usepackage{pifont}
\usepackage{graphicx}
\usepackage{adjustbox}
\usepackage{wrapfig}
\usepackage{makecell}
\usepackage{nicefrac}
\usepackage{microtype}
\usepackage[table,dvipsnames]{xcolor}
\usepackage{amsmath}
\usepackage{multirow}
\usepackage{subcaption}
\usepackage{algorithm}
\usepackage{algorithmic}
\usepackage{float}
\usepackage{dblfloatfix}
\usepackage{tcolorbox}
\usepackage[breaklinks,colorlinks,citecolor=blue,linkcolor=blue,urlcolor=blue]{hyperref}
\usepackage[nameinlink,capitalise,noabbrev]{cleveref}
\tcbuselibrary{skins,breakable}
\newtcolorbox{promptbox}{
  breakable,
  colback=blue!4,
  colframe=blue!45!black,
  boxrule=0.4pt,
  arc=2pt,
  left=5pt,
  right=5pt,
  top=4pt,
  bottom=4pt,
  before skip=2pt,
  after skip=2pt,
}

\definecolor{ncmetabg}{HTML}{F1F4F7}
\definecolor{ncmetaedge}{HTML}{DCE6F5}
\definecolor{ncmetablue}{HTML}{1877F2}
\definecolor{ncapricot}{HTML}{F7EBDD}

\newcommand{\githuburl}{https://github.com/mbzuai-oryx/Ask-Solve-Generate}
\newcommand{\projectpageurl}{https://mbzuai-oryx.github.io/Ask-Solve-Generate/}
\newcommand{\modelsurl}{https://huggingface.co/collections/Ritesh-hf/ask-solve-generate-paper-models}

\newcommand{\githubicon}{\raisebox{-0.18ex}{\makebox[1.35em][c]{\includegraphics[height=1.1em]{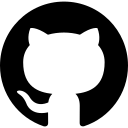}}}}
\newcommand{\projecticon}{\raisebox{-0.18ex}{\makebox[1.35em][c]{\includegraphics[height=1.1em]{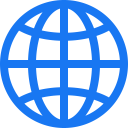}}}}
\newcommand{\hficon}{\raisebox{-0.18ex}{\makebox[1.35em][c]{\includegraphics[height=1.1em]{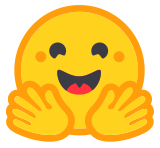}}}}

\newcommand{\cmark}{\ensuremath{\checkmark}}
\newcommand{\xmark}{\ensuremath{\times}}

\newcommand{\dpos}[1]{\textcolor{ForestGreen}{\scriptsize (+#1)}}
\newcommand{\dneg}[1]{\textcolor{BrickRed}{\scriptsize (-#1)}}
\newcommand{\groupcell}[2]{\multirow{#1}{*}{\parbox[c]{1.60cm}{\centering #2}}}
\newcommand{\adopted}[1]{\cellcolor{green!8}#1}

\newif\ifshowcolors
\showcolorsfalse
\ifshowcolors

\else

\fi

\makeatletter
\newcommand{\maketitleboxed}[1]{%
  \par
  \begingroup
    \renewcommand{\thefootnote}{\fnsymbol{footnote}}
    \renewcommand{\@makefnmark}{\hbox to \z@{$^{\@thefnmark}$\hss}}
    \long\def\@makefntext##1{%
      \parindent 1em\noindent
      \hbox to 1.8em{\hss $\m@th ^{\@thefnmark}$}##1%
    }
    \thispagestyle{empty}%
    \begin{tcolorbox}[
      enhanced,
      colback=ncmetabg,
      colframe=ncmetaedge,
      boxrule=0.35pt,
      arc=12pt,
      left=0.55cm, right=0.55cm, top=0.45cm, bottom=0.4cm,
      interior style={shade, shading angle=315,
        left color=white!96!ncmetabg,
        right color=ncmetablue!4!ncapricot!8!ncmetabg},
      before skip=0pt, after skip=0.4em,
      grow to left by=1.5pt, grow to right by=1.5pt,
    ]
      \centering
      {\LARGE\bf \@title\par}%
      \vskip 0.22in
      \def\And{\end{tabular}\hfil\linebreak[0]\hfil\begin{tabular}[t]{c}\bf\rule{\z@}{24\p@}\ignorespaces}%
      \def\AND{\end{tabular}\hfil\linebreak[4]\hfil\begin{tabular}[t]{c}\bf\rule{\z@}{24\p@}\ignorespaces}%
      \begin{tabular}[t]{c}\rule{\z@}{24\p@}\@author\end{tabular}\par
      \vskip 0.18in
      \begingroup
        \leftskip=1.5em \rightskip=1.5em
        \centerline{\large\bf Abstract}\vspace{0.6ex}
        \small #1\par
        \vspace{1.0ex}
        \begingroup
          \leftskip=1.5em \rightskip=1.5em plus 1fil \parfillskip=0pt \parindent=0pt
          \footnotesize
          \noindent\begin{tabular}{@{}c@{\hspace{0.55em}}l@{\hspace{0.35em}}l@{}}
            \githubicon & \textbf{Github Code:} & \href{\githuburl}{\texttt{\githuburl}}\\
            \projecticon & \textbf{Project Page:} & \href{\projectpageurl}{\texttt{\projectpageurl}}\\
            \hficon & \textbf{Models:} & \href{\modelsurl}{\texttt{\modelsurl}}
          \end{tabular}\par
        \endgroup
      \endgroup
    \end{tcolorbox}%
    \@thanks
    \@notice
  \endgroup
  \let\maketitle\relax
  \let\thanks\relax
}
\makeatother

\title{Ask, Solve, Generate: Self-Evolving Unified Multimodal Understanding and Generation via Self-Consistency Rewards}

\author{%
Ritesh Thawkar$^{1}$ \quad
Shravan Venkatraman$^{1}$ \quad
Omkar Thawakar$^{1}$ \quad
Abdelrahman Shaker$^{1}$\\
Fahad Shahbaz Khan$^{1,4}$ \quad
Hisham Cholakkal$^{1}$ \quad
Salman Khan$^{1,3}$ \quad
Rao Muhammad Anwer$^{1,2}$\\[0.35em]
{\normalfont\small $^{1}$Mohamed bin Zayed University of Artificial Intelligence}\\
{\normalfont\small $^{2}$Aalto University \quad
$^{3}$Australian National University \quad
$^{4}$Link\"oping University}%
}

\begin{document}

\maketitleboxed{Most existing unified large multimodal models (LMMs) that jointly perform visual understanding and image generation still rely heavily on curated supervision during post-training, typically requiring human annotations, preference labels, or external reward models.
In this work, we ask: Can a unified LMM autonomously improve both capabilities using only unlabeled images?
To answer this, we propose a self-evolving training framework built around three collaborating internal roles: a Proposer that generates visual questions, a Solver that answers and evaluates them, and a Generator that synthesizes images. Our training is driven by self-derived consistency signals, requiring no human annotations, preference labels, or task-trained external reward/judge model. To stabilize training, we introduce Solver Token Entropy (STE), a continuous difficulty signal derived from token-level prediction uncertainty that remains informative even when sample-level consistency collapses.
For image generation, we further design a multi-scale internal evaluation scheme that combines question--answer fidelity scoring with cycle-consistent captioning. This creates a solver-mediated coupling: as visual understanding improves, the Solver provides more reliable generation-side assessment and stronger internal training signals.
Our framework keeps the same role decomposition, reward logic, and training schedule across diverse unified model paradigms, including diffusion-based (BLIP3o), rectified-flow (BAGEL), and autoregressive (VARGPT-v1.1) architectures, requiring only each backbone's native prompting and generation interface. Across eight reported understanding metrics, our method consistently improves over the respective base unified models. In particular, when applied to BAGEL, our approach achieves an absolute gain of $+3.5\%$ on MMMU and improves image generation performance on GenEval from $82\%$ to $85\%$. Code and models are publicly released.
}

\begin{figure}[!ht]
\centering
\includegraphics[width=\linewidth]{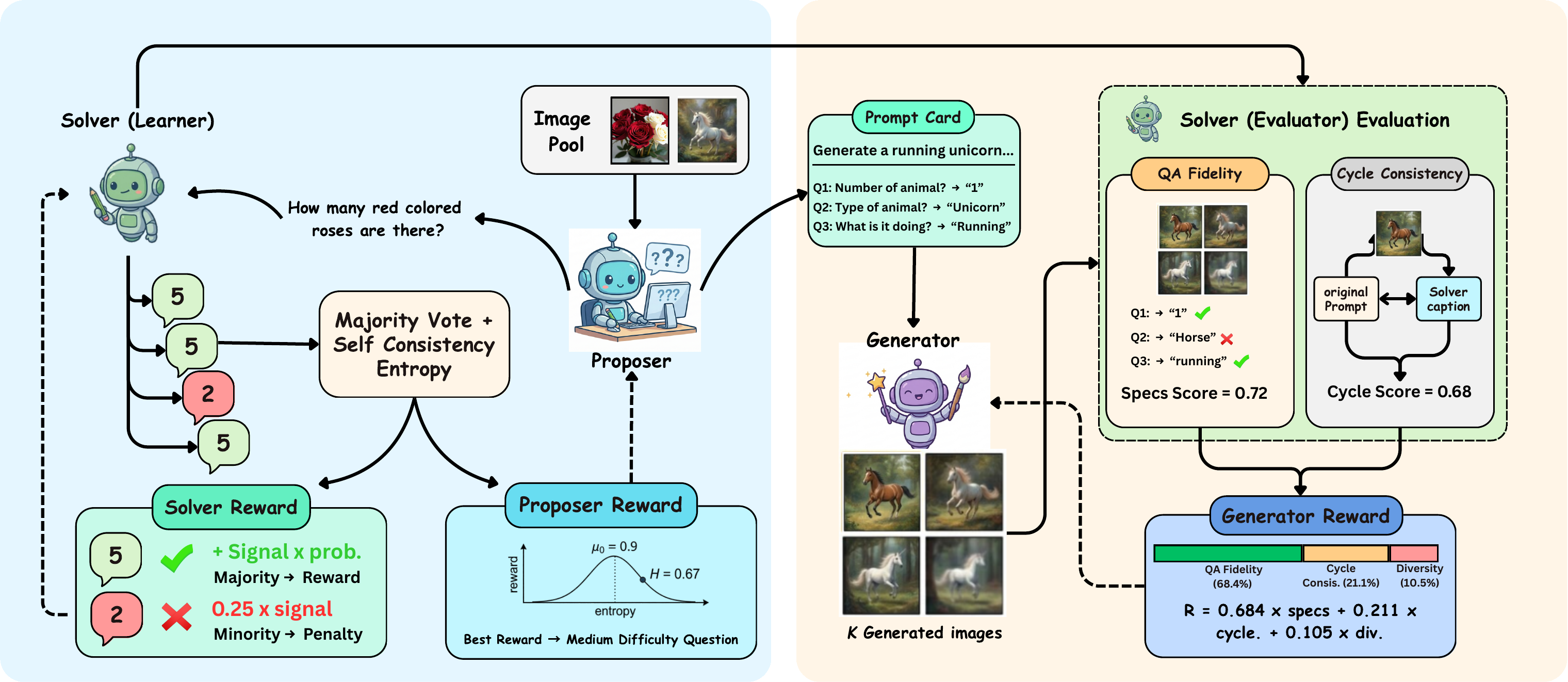}
\caption{\textbf{Overview of our self-evolving framework.} Three LoRA adapters--\emph{Proposer}, \emph{Solver}, and \emph{Generator}--are trained on a frozen backbone using only unlabeled images. The understanding loop uses prompt-perturbed self-consistency and Solver Token Entropy (STE), while the generation loop uses the Solver as an internal evaluator through QA fidelity and cycle-consistent captioning.}
\label{fig:teaser}
\end{figure}

\section{Introduction}

\emph{Unified understanding and generation} models that integrate visual comprehension and image synthesis within one architecture have advanced rapidly~\citep{chen2025januspro, wang2024emu3, qu2025tokenflow, deng2025bagel}. Shared representations create an opportunity for cross-task coupling: stronger understanding can make generation assessment more reliable, while generated visual contexts can expose new perceptual cases. However, pretraining alone does not make this interaction reliable, and substantial headroom remains for post-training improvement on both capabilities.

Most post-training pipelines rely on curated annotations, such as VQA labels for understanding, preference labels for generation, or both, making them costly and brittle. Recent self-improvement methods attempt to reduce this dependency through role-based self-play~\citep{su2025unigame, han2026unicorn}, dual likelihood rewards~\citep{hong2025dualselfrewards}, internal gap exploitation~\citep{han2025internalgap}, and reinforcement-based optimization~\citep{mao2025unirl}. These methods demonstrate strong progress, but their reported validation usually leaves at least one dimension open: one model family, one task direction, or a method-specific signal such as proxy tasks, reconstruction, dual likelihood, self-generated data, or rule rewards. To our knowledge, no prior work reports the same fully self-supervised U+G recipe, using only unlabeled images, with consistent gains across autoregressive, diffusion, and rectified-flow generation paradigms (Table~\ref{tab:qualitative_comparison}).

Removing labels entirely introduces two concrete technical challenges. First, \emph{reward degeneracy}: sample-level self-consistency collapses to zero entropy agreement even when internal confidence is low, producing weak curriculum signals. Second, \emph{weak cross-task coupling}: visual understanding and image generation are often optimized with separate objectives, so generation quality receives little direct benefit from improving comprehension.

To bridge this gap, we propose a self-evolving framework that decomposes a frozen backbone into three LoRA-based roles~\citep{hu2022lora} (Figure~\ref{fig:teaser}): a \textbf{Proposer} that generates visual questions, a \textbf{Solver} that answers and evaluates, and a \textbf{Generator} that synthesizes images. To address reward degeneracy, we introduce \emph{Solver Token Entropy} (STE), a token-level uncertainty signal that remains informative even when sample-level consistency is saturated. To address weak coupling, we use the Solver as an internal evaluator for generation via QA fidelity scoring and cycle-consistent captioning. The resulting interaction is solver-mediated rather than symmetric: Proposer/Solver updates improve the evaluator, and the improved evaluator provides sharper generation rewards. Importantly, the algorithm keeps the role decomposition, reward definitions, and training schedule fixed across backbones, while using only each model's native chat wrapper and generation interface. We validate this backbone-portable recipe across three unified models spanning diffusion (BLIP3o~\citep{chen2025blip3o}), rectified-flow generation (BAGEL~\citep{deng2025bagel}), and autoregressive generation (VARGPT-v1.1~\citep{zhuang2025vargpt}) paradigms. The shared recipe yields consistent within-backbone gains of +1.9 to +3.6 points on the six percentage-based understanding metrics, double-digit gains on MME sub-scores, and +3\% GenEval on all three backbones, providing evidence that one unsupervised self-evolving algorithm can improve both tasks across fundamentally different generation architectures. In summary, our main contributions are:
\begin{itemize}
    \item We propose a self-evolving training framework that improves both visual understanding and image generation capabilities of unified models using only unlabeled images, eliminating the need for human annotations, curated supervision, and task-trained external reward/judge models.
    \item We introduce Solver Token Entropy (STE), a continuous difficulty measure from token-level prediction uncertainty, which resolves the degenerate-signal limitation of sample-level self-consistency and enables effective curriculum learning from a cold start.
    \item We design a multi-scale generation assessment combining QA fidelity scoring with cycle-consistent captioning, using the Solver as an internal evaluator so understanding improvements sharpen generation-side rewards.
    \item We demonstrate backbone-portable generality by applying one self-evolving algorithm, with only backbone-native wrappers, to three unified backbones spanning diffusion, rectified-flow, and autoregressive generation (Table~\ref{tab:qualitative_comparison}).
\end{itemize}



\section{Related Work}
\label{sec:related_work}

\noindent\textbf{Unified understanding and generation models.}
Unified models that jointly perform visual understanding and image generation have advanced rapidly, spanning several increasingly diverse architectural paradigms, including early token fusion~\citep{team2024chameleon}, decoupled vision encoders~\citep{wu2025janus, chen2025januspro}, and token-level unification~\citep{xie2024showo, wu2024vilau}. Recent systems further explore hybrid autoregressive--flow designs~\citep{ma2024janusflow, xie2025showo2}, state-space sequence modeling~\citep{zou2025omnimamba}, and discrete autoregressive multimodal token spaces~\citep{wu2026liquid, geng2025xomni, longcat2026longcatnext}. Modern unified backbones also diversify across diffusion-based~\citep{chen2025blip3o}, rectified-flow~\citep{deng2025bagel}, and autoregressive generation frameworks~\citep{xianwei2025vargpt, zhuang2025vargpt}. Despite these advances, most improvements still rely on supervised post-training with carefully curated paired data, motivating the exploration of fully self-supervised post-training objectives that jointly benefit both visual understanding and image generation.

\noindent\textbf{Role-based self-play and self-improvement.}
Role-based self-play decomposes a unified model into interacting components that generate supervision internally. UniCorn~\citep{han2026unicorn} uses Proposer/Solver/Judge roles to distill latent understanding into generative signals, with reported gains primarily on image generation while maintaining comprehension. UniGame~\citep{su2025unigame} applies a lightweight perturber at the shared token interface and reports improvements on understanding, generation, and robustness within its evaluated setting. EvoLMM~\citep{thawakar2025evolmm} is closest in spirit to our understanding loop, using Proposer/Solver self-consistency to improve multimodal reasoning from raw images, but it targets understanding rather than joint U+G post-training. These methods motivate internal supervision, but they do not report one unchanged self-supervised U+G recipe validated across diffusion, rectified-flow, and autoregressive backbones.

\begin{table}[t]
  \caption{Structured comparison of self-evolving and self-improving methods for unified understanding and generation. We retain the same axes while using \emph{Generalizability} to mean reported validation of the same U+G training recipe across diffusion, rectified-flow, and autoregressive unified backbones. An \xmark{} indicates that this specific cross-paradigm validation is not reported, not that the method is theoretically inapplicable.}
  \label{tab:qualitative_comparison}
  \vspace{0.35em}
  \centering
  \setlength{\aboverulesep}{3pt}
  \setlength{\belowrulesep}{3pt}
  \resizebox{\linewidth}{!}{%
  \begin{tabular}{@{}lcccc@{}}
    \toprule
    \textbf{Method} & \textbf{External Supervision} & \textbf{Tasks} & \textbf{Architecture Specific} & \textbf{Generalizability} \\
    \midrule
    \multicolumn{5}{l}{\emph{Representative prior methods}} \\
    \midrule
    UniGame \cite{su2025unigame}           & Supervised task data    & U+G                 & No (shared-token interface)       & \xmark  \\
    SUDER \cite{hong2025dualselfrewards}   & None                    & U+G                 & No (dual likelihood required)     & \xmark \\
    UniCorn \cite{han2026unicorn}          & None                    & G (+U maintained)   & No                                & \xmark \\
    EvoLMM \cite{thawakar2025evolmm}       & None                    & U only              & No                                & \xmark \\
    SILMM \cite{qu2024silmm}               & None                    & G only              & No (DPO/self-feedback)            & \xmark \\
    RecA \cite{xie2025reconstruction}      & None                    & G/Edit$\leftarrow$U & Yes (hidden embeddings)           & \xmark  \\
    Internal Gap \cite{han2025internalgap} & None                    & G (+U co-improve)   & No                                & \xmark \\
    SRUM \cite{jin2025srum}                & None                    & G$\leftarrow$U      & No (global-local rewards)         & \xmark  \\
    ILLUME \cite{wang2024illume}           & Curated training data   & U+G                 & Yes (tokenizer/training design)   & \xmark \\
    SEER \cite{tang2026endogenous}         & Proxy task (300 samples) & G$\leftarrow$U      & No (reprompting loop)             & \xmark \\
    UniRL \cite{mao2025unirl}              & No external image data      & U+G                 & No (SFT/GRPO recipe)              & \xmark \\
    CoRL \cite{jiang2026coreinforcement}   & Curated task data/rewards & U+G               & No (GRPO recipe)                  & \xmark \\
    UAE \cite{yan2025understandinggeneration} & None (recon. reward) & U+G                 & Yes (auto-encoder coupling)       & \xmark  \\
    \midrule
    \rowcolor{green!8}
    \textbf{Ours}                          & \textbf{None}           & \textbf{U+G}        & \textbf{No}                      & \cmark \\
    \bottomrule
  \end{tabular}%
  }
\end{table}

\noindent\textbf{Reconstruction and internal gap methods.}
Another line couples understanding and generation through reconstruction or by exploiting the common gap where understanding is stronger than generation. UAE~\citep{yan2025understandinggeneration} frames unified models as auto-encoders and optimizes reconstructive rewards between image-to-text understanding and text-to-image generation. RecA~\citep{xie2025reconstruction} uses a model's own visual-understanding embeddings as dense prompts for self-supervised reconstruction, improving generation and editing across several UMM designs, but it requires access to internal embeddings and is not a joint U+G training loop. Internal Gap~\citep{han2025internalgap} uses the understanding module to score generations and construct post-training data, reporting co-improvement, but its supervision is driven by score-based data selection rather than token-level difficulty and multi-scale generation evaluation.

\noindent\textbf{Reinforcement-based optimization.}
RL-based post-training treats both understanding and generation as policy optimization. UniRL~\citep{mao2025unirl} uses self-generated images with SFT/GRPO and reports no external image data, while CoRL~\citep{jiang2026coreinforcement} uses GRPO over curated task data with task/rule rewards; both show that reinforcement-style updates can improve unified models, but their reported recipes remain tied to particular reward designs and backbone families. SUDER~\citep{hong2025dualselfrewards} avoids external reward models through dual likelihood self-rewards. SRUM~\citep{jin2025srum} uses the model's own understanding module as an internal evaluator with global-local rewards, and SILMM~\citep{qu2024silmm} uses self-feedback and DPO for compositional text-to-image generation. ILLUME~\citep{wang2024illume} combines unified next-token modeling, tokenizer/training design, and self-assessment. These methods are complementary, but they do not report the same fully self-supervised U+G recipe across diffusion, rectified-flow, and autoregressive unified backbones.

\noindent\textbf{Our approach.}
Table~\ref{tab:qualitative_comparison} summarizes prior work using the same columns as before, but interprets generalizability as reported evidence of one unchanged U+G training recipe across diffusion, rectified-flow, and autoregressive unified backbones. Our framework uses only unlabeled images and three lightweight LoRA roles on a frozen backbone, deriving training signals from self-consistency, token-level entropy (STE), and multi-scale internal evaluation for generation (QA fidelity + cycle-consistent captioning). Several prior methods are self-supervised or broadly applicable in principle; the distinction is empirical validation under a fixed recipe, not theoretical applicability. To our knowledge, none reports the same fully self-supervised U+G recipe across diffusion, rectified-flow, and autoregressive unified backbones. We provide empirical evidence of consistent gains across diffusion (BLIP3o-8B), rectified-flow (BAGEL), and autoregressive (VARGPT-v1.1) backbones while keeping the role decomposition, reward logic, and schedule fixed and changing only backbone-native wrappers.

\section{Method}
\label{sec:method}

\subsection{Problem Formulation}
\label{sec:problem}

We study \emph{unsupervised} post-training of unified multimodal models, where the only training data are raw images. Let a pretrained unified model $\mathcal{M}_\theta$ expose two native interfaces: (i) visual understanding $\mathcal{U}: (\mathcal{I}, q)\mapsto a$ (answer a question $q$ about image $\mathcal{I}$), and (ii) image generation $\mathcal{G}: t\mapsto \hat{\mathcal{I}}$ (generate an image from text prompt $t$). The training pool is an unlabeled image set $\mathcal{D}=\{\mathcal{I}_1,\ldots,\mathcal{I}_N\}$ with no captions, QA pairs, or preference labels.

To preserve base capabilities and keep the algorithm backbone-portable, we freeze the backbone parameters $\theta$ and train only lightweight role adapters:
\begin{equation}
    \phi=\{\phi_p,\phi_s,\phi_g\}, \qquad \theta'=\theta\oplus\phi,
    \label{eq:adapted_params}
\end{equation}
where $\phi_p$ parameterizes a \textbf{Proposer} $\pi_{\phi_p}(q\mid\mathcal{I})$, $\phi_s$ a \textbf{Solver} $\pi_{\phi_s}(a\mid\mathcal{I},q)$, and $\phi_g$ a \textbf{Generator} $\pi_{\phi_g}(\hat{\mathcal{I}}\mid t)$. We use only model-native operations (QA, captioning, and text-conditioned generation), frozen similarity computations, and each backbone's native trainable generator interface; we do not use external LLM/VLM judges or task-trained reward models.

Our goal is to jointly improve understanding and generation using \emph{self-derived} rewards:
\begin{equation}
    \max_{\phi_p,\phi_s,\phi_g}\ \mathcal{J}(\phi)=
    \lambda_u\,\mathbb{E}_{\tau_u\sim\pi_{\phi_p,\phi_s}}\!\left[R_p(\tau_u)+R_s(\tau_u)\right]
    +\lambda_g\,\mathbb{E}_{\tau_g\sim\pi_{\phi_p,\phi_s,\phi_g}}\!\left[R_g(\tau_g)\right],
    \label{eq:joint_objective}
\end{equation}
subject to: (C1) no curated supervision, (C2) no task-trained external reward/judge model, and (C3) controlled policy drift via KL regularization to frozen reference policies. Eq.~\eqref{eq:joint_objective} naturally separates into two loops: an \emph{understanding loop} (Proposer/Solver) and a \emph{generation loop} (Generator evaluated by the Solver). Below we define the rewards $R_p$, $R_s$, and $R_g$.

\subsection{Framework Overview}
\label{sec:overview}

\begin{figure*}[t]
  \centering
  \includegraphics[width=\linewidth]{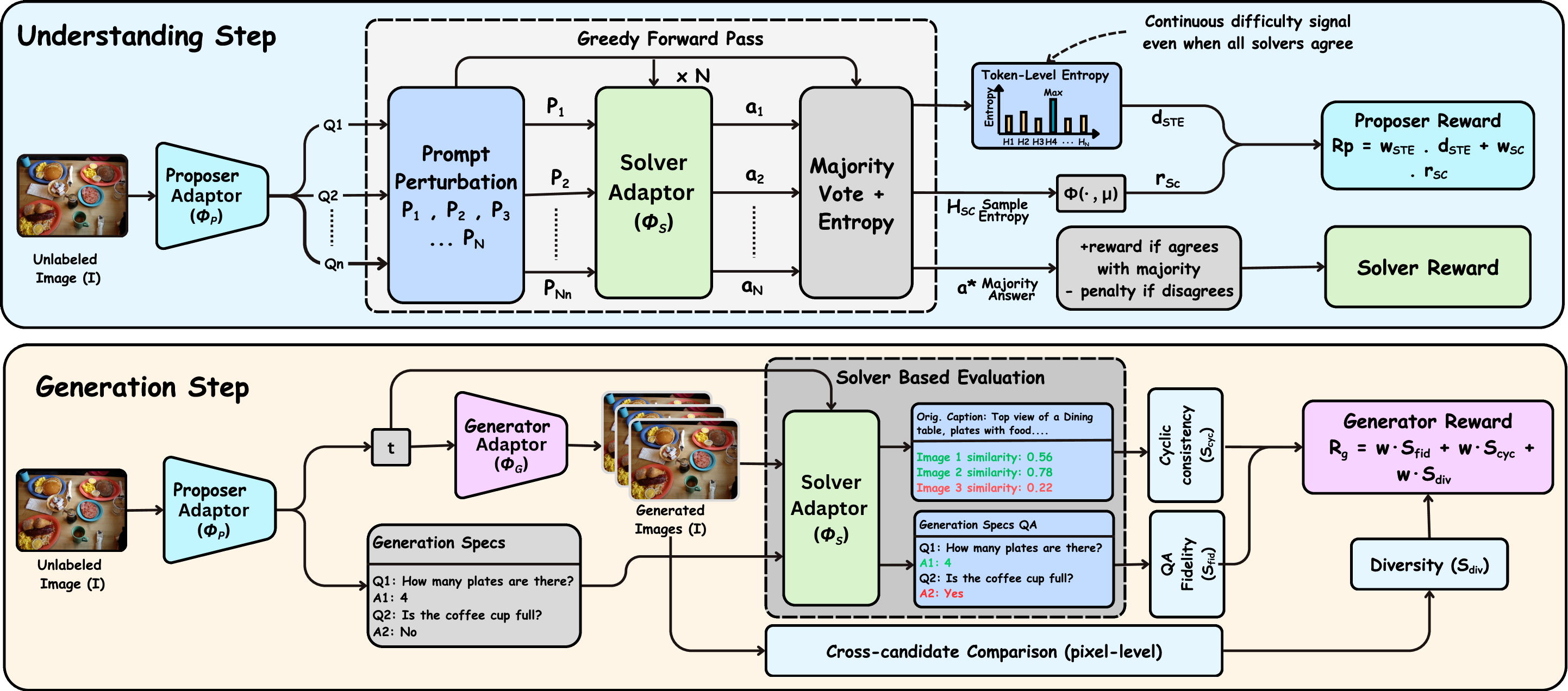}
  \caption{Overview of our \emph{Proposer}--\emph{Solver}--\emph{Generator} self-evolving framework. Given only a frozen backbone and unlabeled images, we attach three lightweight LoRA adapters for the \emph{Proposer}, \emph{Solver}, and \emph{Generator} roles. In understanding steps (left), the \emph{Proposer} generates visual questions, and the \emph{Solver} answers under multiple prompt perturbations; self-consistency agreement and Solver Token Entropy (STE) jointly produce the training signal, encouraging informative questions at the \emph{Solver}'s competence frontier. In generation steps (right), the \emph{Generator} synthesizes images from prompt cards and the same \emph{Solver} evaluates them via QA fidelity and cycle-consistent captioning. Thus, visual-understanding updates improve the internal evaluator that supplies generation rewards, while all roles remain trained without labels or task-trained reward/judge models.}
  \label{fig:framework}
\end{figure*}

We instantiate all three roles with LoRA adapters~\cite{hu2022lora} on the same frozen backbone (Figure~\ref{fig:framework}). Training alternates between \emph{understanding steps}, which update the Proposer and Solver from self-consistency-based rewards, and \emph{generation steps}, which update the Generator using the Solver as an internal evaluator. The coupling is solver-mediated rather than a symmetric gradient exchange: Proposer/Solver updates improve the evaluator, and the improved evaluator supplies sharper generation rewards.

\subsection{Self Consistent Understanding}
\label{sec:understanding}

We learn to ask and answer informative questions from unlabeled images using two complementary signals: (i) \emph{framing-level robustness}, which measures whether answers remain stable under equivalent instructions, and (ii) \emph{token uncertainty}, which remains informative when output-level agreement becomes degenerate.

For an image $\mathcal{I}$, the Proposer samples a question $q \sim \pi_{\phi_p}(\cdot \mid \mathcal{I})$. We query the Solver under $N$ fixed prompt framings $\{\rho_1,\ldots,\rho_N\}$ (templates that preserve semantics but vary wording), i.e., $a_i=\text{Solver}_{\phi_s}(\mathcal{I}, \rho_i(q))$ for $i=1,\ldots,N$.
Let $p_c$ be the empirical distribution over distinct answers, and $a^*$ the modal answer. We measure sample-level self-consistency by entropy:
\begin{equation}
    H_{\text{sc}} = -\sum_{c} p_c \ln p_c, \quad p_c = \frac{|\{i : a_i = c\}|}{N}.
    \label{eq:sample_entropy}
\end{equation}
Low $H_{\text{sc}}$ indicates robustness to rephrasing; high $H_{\text{sc}}$ indicates unstable reasoning.

\paragraph{Solver Token Entropy (STE).}
$H_{\text{sc}}$ can collapse (all framings yield the same answer) even when the model is uncertain, producing weak curriculum signals. To recover a continuous notion of difficulty, we compute a token-level uncertainty score from the Solver's next-token distributions. Let $\mathbf{p}_t$ be the next-token distribution at decoding step $t$ of an answer with length $T$, and $H_t = -\sum_{v} p_{t,v} \ln p_{t,v}$ its entropy. In implementation we evaluate the first $K_{\text{ste}}{=}5$ answer tokens and use the maximum entropy:
\begin{equation}
    \hat{H} = \max_{t \in \{1, \ldots, \min(T,K_{\text{ste}})\}} H_t,
    \label{eq:ste}
\end{equation}
and convert $\hat{H}$ into a stable difficulty score $d_{\text{ste}}\in[0,1]$ by taking its percentile rank within a rolling window of recent $\hat{H}$ values. The max operator targets decisive uncertain content tokens in short answers (e.g., count, color, relation, or negation), while the rolling-window normalization makes STE comparable across training as the Solver's confidence distribution shifts. STE is a scalar uncertainty score from the Solver; it is not a cross-prompt token-probability matching objective.

\paragraph{Proposer and Solver optimization.}
The Proposer is trained to generate questions near the Solver's competence frontier: questions that are neither trivial nor unsolvable. To this end, we compute a bounded self-consistency score $r_{\text{sc}}\in[-1,1]$ that combines (i) an \emph{adaptive target} favoring medium self-consistency entropy (a Gaussian centered at a running EMA target $\mu$, following curriculum and automated curriculum-learning principles~\citep{bengio2009curriculum, graves2017automated}) and (ii) a \emph{local informativeness} term that down-weights degenerate cases such as unanimous votes with large margins. To prevent reward degeneracy when $H_{\text{sc}}\!\approx\!0$, we add STE difficulty:
\begin{equation}
    R_p = w_{\text{ste}} \cdot d_{\text{ste}} + w_{\text{sc}} \cdot r_{\text{sc}},
    \label{eq:proposer_reward}
\end{equation}
which yields an automatic curriculum that tracks the Solver as it improves.

The Solver is trained to answer consistently across framings by rewarding agreement with the modal answer $a^*$, while penalizing non-canonical or low-information outputs and excessive verbosity. In practice, we optimize both Proposer and Solver with KL-regularized policy gradients and EMA baselines (GRPO over groups of $K$ sampled questions for the Proposer; REINFORCE over answer tokens for the Solver), using an adaptive KL coefficient to control drift from the frozen reference model. This design keeps the optimization stable while allowing the Proposer--Solver pair to co-evolve under purely internal signals.

\subsection{Generation Assessment via Internal Evaluation}
\label{sec:generation}

We improve generation by treating the Solver as an internal evaluator. Given a prompt $t$ for a source image $\mathcal{I}$ (obtained by having the Solver caption $\mathcal{I}$), the Generator produces a candidate image $\hat{\mathcal{I}}=\text{Generator}_{\phi_g}(t)$. We compute two complementary rewards.

\paragraph{QA fidelity scoring.}
We sample $M$ diagnostic questions about $\mathcal{I}$ (from the Proposer) and obtain Solver-derived reference answers $a_j^{\text{ref}}$ by querying the same Solver on $\mathcal{I}$. These reference answers are generated internally from the real/source image; they are not Proposer labels, human annotations, dataset labels, or external supervision. Low-quality prompt cards are filtered by spec-quality and minimum-QA gates before any generator update. We then ask the same questions on the generated image:
\begin{equation}
    S_{\text{fid}} = \frac{1}{M} \sum_{j=1}^{M} m(\hat{a}_j, a^{\text{ref}}_j),
    \label{eq:fidelity}
\end{equation}
where $m(\cdot,\cdot)$ is a soft answer match combining exact/substring/numeric agreement with the Solver majority fraction. $S_{\text{fid}}$ measures how well local attributes, counts, and relations are preserved from $\mathcal{I}$ to $\hat{\mathcal{I}}$.

\paragraph{Cycle consistent captioning.}
QA fidelity is local; we add a global check by captioning the generated image to obtain $\hat{t}$ and measuring cycle consistency:
\begin{equation}
    S_{\text{cyc}} = \tfrac{1}{2}\,\text{sim}(f(t), f(\hat{t})) + \tfrac{1}{2}\,\text{sim}_{\text{vl}}(\hat{\mathcal{I}}, t),
    \label{eq:cycle}
\end{equation}
where $f(\cdot)$ is a frozen text embedding and $\text{sim}_{\text{vl}}$ is a frozen vision--language similarity.
In practice these are fixed similarity backends (model-native embeddings where available, and CLIP-style image-text similarity in wrappers that require it), with lexical overlap used only as a fallback for text-caption similarity. They are not trained reward models.

\paragraph{Generator optimization.}
The Generator reward combines these signals, plus small diversity/contradiction terms. $S_{\text{div}}$ is a leave-one-out diversity contribution across the candidate images for the same prompt, and $S_{\text{ctr}}$ penalizes explicit yes/no polarity conflicts between expected and Solver-predicted answers:
\begin{equation}
    R_g = w_{\text{fid}} \cdot S_{\text{fid}} + w_{\text{cyc}} \cdot S_{\text{cyc}} + w_{\text{div}} \cdot S_{\text{div}} - w_{\text{ctr}} \cdot S_{\text{ctr}},
    \label{eq:gen_reward}
\end{equation}
We keep the reward definition and trainable-adapter constraint fixed across backbones, but route the update through each Generator's native parameterization: autoregressive generators use token-policy updates over image-token traces, while diffusion/flow generators use reward-weighted denoising updates to the generator-side LoRA modules. Thus the supervision signal is shared, while the low-level update follows the backbone's generation interface.

\paragraph{Coupling design.}
The two loops are coupled only through the Solver: the Proposer/Solver are trained from understanding signals, while generation rewards become stronger as the Solver becomes a more reliable evaluator. This makes the direct mechanism strongest from understanding to generation; the generation loop contributes indirectly by exposing the Solver to diverse generated visual contexts and by keeping Generator updates aligned with the evolving evaluator.

\section{Experiments}
\label{sec:experiments}

\subsection{Experimental Setup}
\label{sec:setup}

We apply one training recipe to three unified backbones spanning the major image synthesis paradigms: BLIP3o-8B~\cite{chen2025blip3o} (diffusion), BAGEL~\cite{deng2025bagel} (rectified flow) and VARGPT-v1.1~\cite{zhuang2025vargpt} (autoregressive). For each backbone, we attach three LoRA adapters~\cite{hu2022lora} for the Proposer, Solver, and Generator roles while keeping the base model frozen. Training uses only unlabeled images from a 10{,}000 image pool sampled from five open datasets (COCO, SA-1B, TextVQA, GQA, and LAION-COCO); all annotations, boxes, captions, and QA labels are discarded, so TextVQA/GQA overlap is image-only and distributional rather than supervised. Our primary evidence is within-backbone base-versus-ours deltas under identical inference settings. Visual understanding is evaluated on seven benchmarks: MMMU~\cite{yue2024mmmu}, MMBench~\cite{liu2024mmbench}, TextVQA~\cite{singh2019textvqa}, SEED-Bench~\cite{li2024seedbench}, RealWorldQA~\cite{xai2024realworldqa}, MM-Vet~\cite{yu2023mmvet}, and MME~\cite{fu2024mme} (reporting both perception and cognition sub-scores); image generation on GenEval~\cite{ghosh2024geneval}. Prior method scores are reported as context only.

Unless stated otherwise, all runs share the same hyperparameters: a roughly 10k-step training horizon with AdamW in bfloat16, learning rate $1\times10^{-6}$, LoRA rank 16 ($\alpha{=}32$, dropout $0.05$), and a 3:2 understanding to generation schedule. Self-consistency uses $N{=}7$ prompt perturbations, Proposer samples $K{=}3$ candidate questions per image, and STE uses a rolling window of 128 samples.

\subsection{Results and Analysis}
\label{sec:main_results}

\begin{table*}[t]
  \caption{Visual understanding results across unified models and self-evolving methods. We apply the same algorithmic recipe to three backbones, keeping data, role decomposition, reward design, and schedule fixed while using only each backbone's native wrappers. Our method consistently improves all eight reported metrics over each base checkpoint, e.g., MMMU $50.6{\to}52.8$ on BLIP3o-8B, $55.3{\to}58.8$ on BAGEL, and $48.6{\to}51.6$ on VARGPT-v1.1. Reasoning-heavy metrics show the largest gains (MMMU up to +3.5, MME cognition up to +14.9), as well as perceptual metrics (MMBench, TextVQA), confirming that self-evolving signals sharpen reasoning without degrading recognition. These improvements emerge from a 10k-image unlabeled pool and roughly 10k training steps, without human annotations, curated supervision, or task-trained external reward/judge models.}
  \label{tab:understanding_results}
  \centering
  \small

  \resizebox{\linewidth}{!}{%
  \begin{tabular}{@{}llccccccccc@{}}
    \toprule
    \textbf{Method} & \textbf{Baseline} & \textbf{Params} & \textbf{MMMU}  & \textbf{MMBench} & \textbf{TextVQA} & \textbf{SEED} & \textbf{RWQA} & \textbf{MMVet} & \textbf{MME-P} & \textbf{MME-C} \\
    \midrule
    \multicolumn{11}{c}{\emph{Unified understanding and generation models}} \\
    \midrule
    \noalign{\vskip 1.5pt}
    Chameleon \cite{team2024chameleon} & Chameleon & -- & 22.4 & 19.8 & -- & 27.2 & 39.0 & 8.3 & 202.7 & -- \\
    Show-o \cite{xie2024showo} & Show-o & -- & 27.4 & -- & -- & -- & -- & -- & 1232.9 & -- \\
    VILA-U \cite{wu2024vilau} & VILA-U & -- & -- & -- & 60.8 & 59.0 & -- & 33.5 & 1401.8 & -- \\
    Janus \cite{wu2025janus} & Janus & -- & 30.5 & 69.4 & -- & 63.7 & -- & 34.3 & 1338.0 & -- \\
    Janus-Pro-7B \cite{chen2025januspro} & Janus-Pro-7B & 7B & 41.0 & 79.2 & -- & 72.1 & -- & 50.0 & 1567.1 & -- \\
    SEED-X \cite{ge2024seedx} & SEED-X & -- & 35.6 & 70.1 & -- & 66.5 & -- & 43.0 & 1457.0 & -- \\
    Emu3 \cite{wang2024emu3} & Emu3 & -- & 31.6 & 58.5 & 64.7 & 68.2 & 57.4 & 37.2 & 1243.8 & 266.1 \\
    TokenFlow \cite{qu2025tokenflow} & TokenFlow & -- & 43.2 & 76.8 & 62.3 & 72.6 & 56.6 & 48.2 & 1551.1 & 371.1 \\
    MetaMorph \cite{tong2024metamorph} & MetaMorph & -- & 41.8 & 75.2 & 60.5 & 71.8 & 58.3 & -- & -- & -- \\
    \midrule
    \multicolumn{11}{c}{\emph{Self-evolving methods}} \\
    \midrule
    \noalign{\vskip 1.5pt}
    UniGame \cite{su2025unigame} & Janus-Pro-7B & 7B & 43.8 & 83.2 & -- & -- & -- & -- & -- & -- \\
    SUDER \cite{hong2025dualselfrewards} & Janus-Pro-7B & 7B & -- & 80.1 & -- & 71.9 & -- & -- & -- & -- \\
    UniCorn \cite{han2026unicorn} & BAGEL & 7B active & 53.8 & 84.1 & -- & -- & -- & -- & 1660.0 & 677.0 \\
    ILLUME \cite{wang2024illume} & Vicuna-7B & 7B & 38.2 & 75.1 & 72.1 & 72.9 & -- & 37.0 & 1445.3 & -- \\
    \midrule
    BLIP3o-8B \cite{chen2025blip3o} & BLIP3o-8B & 8B & 50.6 & 83.5 & 83.1 & 77.5 & 69.0 & 66.6 & 1682.6 & 647.1 \\
    \rowcolor{green!8}
    BLIP3o-8B (Ours) & BLIP3o-8B & 8B & \textbf{52.8} \dpos{2.2} & \textbf{86.1} \dpos{2.6} & \textbf{85.2} \dpos{2.1} & \textbf{79.4} \dpos{1.9} & \textbf{70.9} \dpos{1.9} & \textbf{68.7} \dpos{2.1} & \textbf{1698.4} \dpos{15.8} & \textbf{660.3} \dpos{13.2} \\
    BAGEL \cite{deng2025bagel} & BAGEL & 7B active & 55.3 & 85.0 & 86.0 & 79.3 & 71.2 & 67.2 & 1687.0 & 701.0 \\
    \rowcolor{green!8}
    BAGEL (Ours) & BAGEL & 7B active & \textbf{58.8} \dpos{3.5} & \textbf{87.1} \dpos{2.1} & \textbf{88.5} \dpos{2.5} & \textbf{81.8} \dpos{2.5} & \textbf{73.9} \dpos{2.7} & \textbf{69.5} \dpos{2.3} & \textbf{1701.7} \dpos{14.7} & \textbf{715.9} \dpos{14.9} \\
    VARGPT-v1.1 \cite{zhuang2025vargpt} & VARGPT-v1.1 & 7B+2B & 48.6 & 81.0 & 82.0 & 76.1 & 67.5 & 51.9 & 1678.3 & 592.9 \\
    \rowcolor{green!8}
    VARGPT-v1.1 (Ours) & VARGPT-v1.1 & 7B+2B & \textbf{51.6} \dpos{3.0} & \textbf{83.7} \dpos{2.7} & \textbf{84.8} \dpos{2.8} & \textbf{79.2} \dpos{3.1} & \textbf{71.1} \dpos{3.6} & \textbf{54.0} \dpos{2.1} & \textbf{1695.7} \dpos{17.4} & \textbf{606.4} \dpos{13.5} \\
    \bottomrule
  \end{tabular}%
  }
\end{table*}

\begin{table*}[t]
\caption{Image generation results on GenEval across unified models and self-evolving methods. We apply the same algorithmic recipe to three backbones and report six compositional subcategories alongside the overall score. Our method yields a uniform +3 percentage-point overall improvement on all three backbones ($84\%{\to}87\%$ on BLIP3o-8B, $82\%{\to}85\%$ on BAGEL, $53\%{\to}56\%$ on VARGPT-v1.1), with the largest gains concentrated on composition-heavy subcategories: Two Objects $85\%{\to}93\%$ and Counting $63\%{\to}71\%$ on BLIP3o-8B, while already-saturated Single Object (${\geq}96\%$) remains stable. Despite very different absolute baselines, the generation loop adapts to each backbone's current capability without changing the reward design or training schedule.}
  \label{tab:generation_results}
  \centering
  \small

  \resizebox{\linewidth}{!}{%
  \begin{tabular}{@{}lllccccccc@{}}
    \toprule
    \multirow{2}{*}{\textbf{Method}} & \multirow{2}{*}{\textbf{Baseline}} & \multirow{2}{*}{\textbf{Params}} & \multicolumn{7}{c}{\textbf{GenEval}} \\
    \cmidrule(lr){4-10}
     &  &  & \textbf{Single Obj.} & \textbf{Two Obj.} & \textbf{Counting} & \textbf{Colors} & \textbf{Position} & \textbf{Color Attri.} & \textbf{Overall} \\
    \midrule
    \multicolumn{10}{c}{\emph{Unified understanding and generation models}} \\
    \midrule
    \noalign{\vskip 1.5pt}
    Chameleon \cite{team2024chameleon} & Chameleon & -- & -- & -- & -- & -- & -- & -- & 39 \\
    Show-o \cite{xie2024showo} & Show-o & -- & 98 & 85 & 67 & 81 & 28 & 55 & 69 \\
    Janus \cite{wu2025janus} & Janus & -- & 97 & 68 & 30 & 84 & 46 & 42 & 61 \\
    Janus-Pro-7B \cite{chen2025januspro} & Janus-Pro-7B & 7B & 99 & 89 & 59 & 90 & 79 & 66 & 80 \\
    SEED-X \cite{ge2024seedx} & SEED-X & -- & 97 & 58 & 26 & 80 & 19 & 14 & 49 \\
    Emu3 \cite{wang2024emu3} & Emu3 & -- & 98 & 71 & 34 & 81 & 17 & 21 & 54 \\
    TokenFlow \cite{qu2025tokenflow} & TokenFlow & -- & 97 & 66 & 40 & 84 & 17 & 26 & 55 \\
    \midrule
    \multicolumn{10}{c}{\emph{Self-evolving methods}} \\
    \midrule
    \noalign{\vskip 1.5pt}
    UniGame \cite{su2025unigame} & Janus-Pro-7B & 7B & 99 & 91 & 62 & 93 & 80 & 68 & 82 \\
    SUDER \cite{hong2025dualselfrewards} & Janus-Pro-7B & 7B & 99 & 89 & 70 & 92 & 82 & 71 & 84 \\
    UniRL (SFT) \cite{mao2025unirl} & Show-o & -- & 99 & 93 & 62 & 89 & 55 & 68 & 77 \\
    UniCorn \cite{han2026unicorn} & BAGEL & 7B active & 99 & 94 & 80 & 88 & 61 & 73 & 82 \\
    ILLUME \cite{wang2024illume} & Vicuna-7B & 7B & 99 & 86 & 45 & 71 & 39 & 28 & 61 \\
    \midrule
    BLIP3o-8B \cite{chen2025blip3o} & BLIP3o-8B & 8B & 100 & 85 & 63 & 92 & 90 & 74 & 84 \\
    \rowcolor{green!8}
    BLIP3o-8B (Ours) & BLIP3o-8B & 8B & 99 \dneg{1\%} & 93 \dpos{8\%} & 71 \dpos{8\%} & 94 \dpos{2\%} & 90 \dpos{0\%} & 75 \dpos{1\%} & \textbf{87} \dpos{3\%} \\
    BAGEL \cite{deng2025bagel} & BAGEL & 7B active & 99 & 94 & 81 & 88 & 64 & 63 & 82 \\
    \rowcolor{green!8}
    BAGEL (Ours) & BAGEL & 7B active & 99 \dpos{0\%} & 95 \dpos{1\%} & 87 \dpos{6\%} & 90 \dpos{2\%} & 67 \dpos{3\%} & 72 \dpos{9\%} & \textbf{85} \dpos{3\%} \\
    VARGPT-v1.1 \cite{zhuang2025vargpt} & VARGPT-v1.1 & 7B+2B & 96 & 53 & 48 & 83 & 13 & 21 & 53 \\
    \rowcolor{green!8}
    VARGPT-v1.1 (Ours) & VARGPT-v1.1 & 7B+2B & 97 \dpos{1\%} & 59 \dpos{6\%} & 56 \dpos{8\%} & 85 \dpos{2\%} & 15 \dpos{2\%} & 24 \dpos{3\%} & \textbf{56} \dpos{3\%} \\
    \bottomrule
  \end{tabular}%
  }
\end{table*}

\noindent\textbf{Base model landscape.} As shown in Tables~\ref{tab:understanding_results} and~\ref{tab:generation_results}, all three backbones achieve strong perceptual performance (MMBench ${\geq}81.0$, TextVQA ${\geq}82.0$) but exhibit characteristic weaknesses: BLIP3o-8B and BAGEL lag in multi-discipline reasoning (MMMU: 50.6 and 55.3), VARGPT-v1.1 in open-ended analysis (MM-Vet: 51.9). For generation, a single object is near saturation (${\geq}96\%$) but compositional subcategories vary widely (e.g. position: 90\% on BLIP3o-8B vs.\ 13\% on VARGPT-v1.1). The framework must therefore improve reasoning and composition without degrading perception.

\noindent\textbf{Visual understanding improves consistently, with the largest gains on the weakest capabilities.} The largest relative gains appear on each backbone's weakest percentage-based metric: MMMU, where all three models start lowest and improve by up to +3.5. Other reasoning-heavy metrics (MM-Vet, MME cognition) follow the same pattern, while perceptual metrics closer to each backbone's ceiling (MMBench, TextVQA) show smaller but consistent improvements, confirming that self-evolving signals sharpen reasoning without degrading recognition.

\noindent\textbf{Image Generation gains concentrate on the hardest compositional subcategories.} The uniform +3\% overall gain masks distinct backbone-specific profiles. On BLIP3o-8B, gains concentrate on the weakest subcategories (Two Objects, Counting) while near-saturated scores are effectively unchanged. On BAGEL, gains are strongest in Color Attribution and Counting, with smaller improvements in Position, Colors, and Two Objects. On VARGPT-v1.1, Counting and Two Objects improve most, while Position ($13{\to}15\%$) and Color Attribution ($21{\to}24\%$) show small gains but remain low overall, reflecting autoregressive limitations that semantic-level supervision cannot fully overcome.

\noindent\textbf{Cross-method comparison and overall pattern.} Among prior self-supervised methods, UniCorn~\cite{han2026unicorn} reports 53.8 MMMU and 82\% GenEval, while SUDER~\cite{hong2025dualselfrewards} achieves 80.1 MMBench and 84\% GenEval. Our BAGEL results (58.8 MMMU, 87.1 MMBench, 85\% GenEval) are competitive with or exceed these without external LLM/VLM judges, curated supervision, or architecture-specific modules; we emphasize that cross-paper comparisons remain unreliable and that the primary evidence is the consistent base-versus-ours deltas under matched conditions.

\subsection{Ablation Study}
\label{sec:ablation}

\begin{figure}[!htbp]
  \centering
  \includegraphics[width=\linewidth]{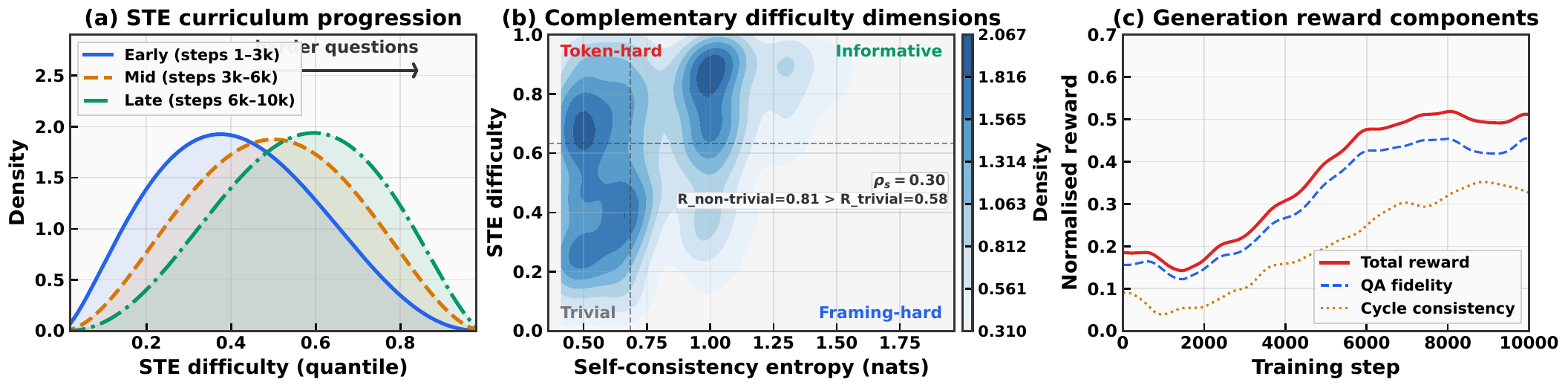}
  \caption{Signal analysis on BLIP3o-8B revealing the complementary roles of our self-evolving training signals. (a)~The STE difficulty distribution shifts toward harder quantiles as training progresses, reflecting the emergence of an adaptive curriculum where the \emph{Proposer} generates increasingly challenging questions matched to the \emph{Solver}'s evolving competence. (b)~Self-consistency entropy and STE occupy complementary regions of the difficulty space: self-consistency captures framing-level robustness while STE detects token-level uncertainty, together providing a richer supervision signal than either alone. (c)~Both QA fidelity and cycle-consistency generation rewards increase monotonically, showing that the \emph{Solver}'s improving understanding is associated with more discriminative generation evaluation.}
  \label{fig:signal_analysis}
\end{figure}

In Table~\ref{tab:ablation}, we show ablations over BLIP3o-8B with the same compute budget as the full method. The two understanding signals are complementary, each capturing distinct failure modes; replacing prompt perturbations with temperature sampling also underperforms, confirming that framing variation provides a stronger robustness signal than stochastic decoding. The supplementary sweeps further show that the STE rolling window is stable over nearby values ($W{=}64,128,256$). On generation, QA fidelity contributes more than cycle-consistency, but both are needed for best joint performance. Figure~\ref{fig:signal_analysis} corroborates these findings: STE shifts toward harder questions over training, the two signals occupy complementary difficulty regions, and both generation rewards increase monotonically.

\begin{table}[!htbp]
  \caption{Ablation study on BLIP3o-8B isolating the contribution of each framework component under a matched training budget. Each row removes or replaces one component while keeping all others at their default configuration. Removing STE reduces MMMU by $-1.8$ and MMBench by $-1.6$, while removing self-consistency reduces MMMU by $-1.3$ and SEED by $-1.1$, showing that both understanding signals are complementary. On the generation side, removing QA fidelity causes the largest GenEval drop ($87\%{\to}84\%$), while removing cycle consistency also degrades compositional quality ($87\%{\to}85\%$). The full framework achieves the best joint result across both tasks, confirming that all components contribute non-redundant supervision.}
  \label{tab:ablation}
  \centering
  \setlength{\tabcolsep}{3pt}
  \resizebox{\linewidth}{!}{%
  \begin{tabular}{@{}lccccccccc@{}}
    \toprule
    & \multicolumn{8}{c}{\textbf{Understanding}} & \textbf{Gen.} \\
    \cmidrule(l){2-10}
    \textbf{Configuration} & \textbf{MMMU} & \textbf{MMB} & \textbf{TVQA} & \textbf{SEED} & \textbf{RWQA} & \textbf{MMVet} & \textbf{MME-P} & \textbf{MME-C} & \textbf{GenEval} \\
    \midrule
    BLIP3o-8B (base) & 50.6 & 83.5 & 83.1 & 77.5 & 69.0 & 66.6 & 1682.6 & 647.1 & 84 \\
    \rowcolor{green!8}
    Full framework (Ours) & \textbf{52.8} & \textbf{86.1} & \textbf{85.2} & \textbf{79.4} & \textbf{70.9} & \textbf{68.7} & \textbf{1698.4} & \textbf{660.3} & \textbf{87} \\
    \midrule
    \midrule
    \noalign{\vskip 1.5pt}
    \quad w/o STE (self-consistency only) & 51.0 & 84.5 & 83.8 & 77.9 & 69.5 & 67.1 & 1690.2 & 652.0 & 86 \\
    \quad w/o self-consistency (STE only) & 51.5 & 85.0 & 84.2 & 78.3 & 70.0 & 67.5 & 1693.5 & 655.1 & 86 \\
    \quad w/o prompt perturbation (temp.\ sampling) & 52.0 & 85.5 & 84.5 & 78.8 & 70.2 & 68.0 & 1695.0 & 657.0 & 86 \\
    \midrule
    \midrule
    \noalign{\vskip 1.5pt}
    \quad w/o QA fidelity scoring & 52.5 & 85.8 & 84.9 & 79.1 & 70.7 & 68.4 & 1697.0 & 658.5 & 84 \\
    \quad w/o cycle-consistent captioning & 52.3 & 85.6 & 84.7 & 79.0 & 70.5 & 68.2 & 1696.5 & 658.0 & 85 \\
    \bottomrule
  \end{tabular}%
  }
\end{table}

\begin{figure}[!htbp]
  \centering
  \includegraphics[width=\linewidth]{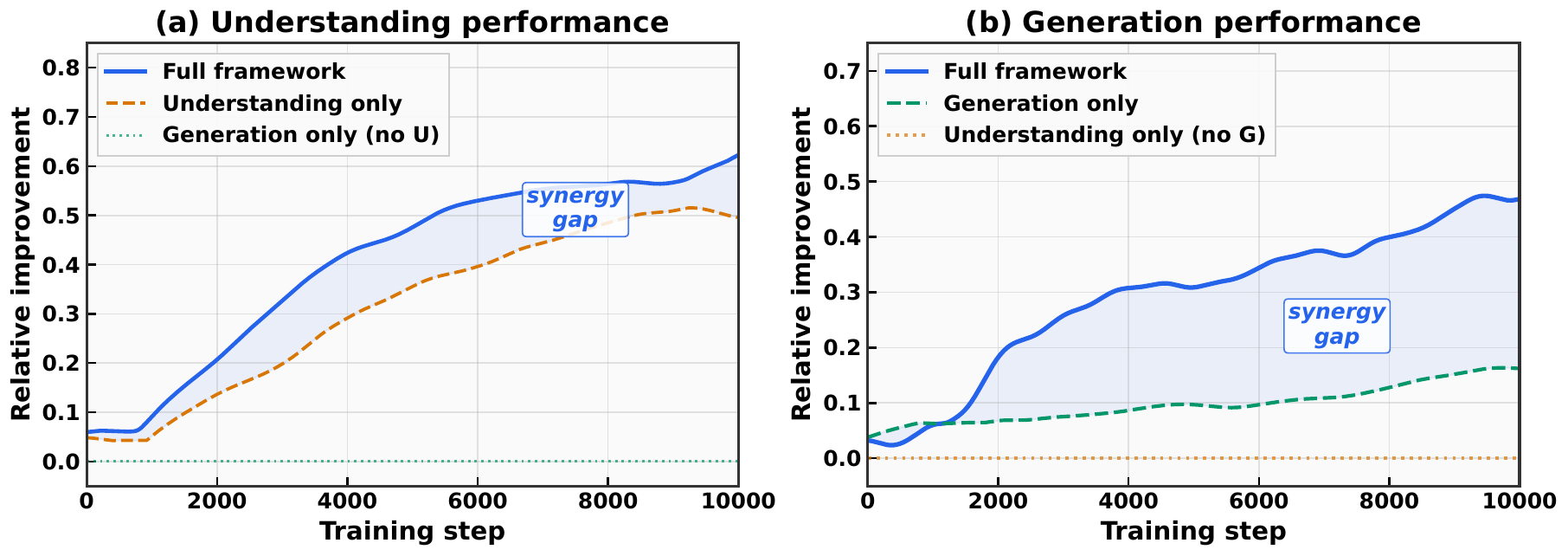}
  \caption{Solver-mediated loop dynamics on BLIP3o-8B comparing joint training against single-loop variants over a roughly 10k-step horizon. (a)~For visual understanding, the full framework (blue) outperforms understanding-only training (orange), while generation-only training (green) stays at the base understanding level because it does not update the Proposer/Solver path. (b)~For image generation, the full framework exceeds generation-only training, consistent with improved visual understanding yielding a more discriminative internal evaluator. The gap should be interpreted as solver-mediated asymmetric coupling, not symmetric gradient exchange between loops.}
  \label{fig:loop_coupling}
\end{figure}

\noindent\textbf{Loop Coupling Analysis.} Table~\ref{tab:loop_ablation} and Figure~\ref{fig:loop_coupling} show that joint training exceeds both single-loop variants, while training on self-generated images only also underperforms. This supports an asymmetric solver-mediated mechanism: understanding improves the Solver's generation evaluations, while generation-only updates improve image synthesis without directly training the Proposer/Solver path.

\begin{table}[!htbp]
  \caption{Loop coupling analysis on BLIP3o-8B comparing the full framework against single-loop variants. Understanding-only training improves MMMU ($50.6\%{\to}52.5\%$) but slightly degrades generation ($84\%{\to}83\%$). Generation-only training reaches $86\%$ GenEval while Solver-side understanding remains at the base checkpoint ($50.6\%$ MMMU). The full framework surpasses both single-loop configurations on both tasks ($52.8\%$ MMMU, $87\%$ GenEval), supporting solver-mediated asymmetric coupling: a stronger \emph{Solver} supplies better generation rewards, while Generator-only updates do not directly improve Solver evaluations.}
  \label{tab:loop_ablation}
  \centering
  \setlength{\tabcolsep}{3pt}
  \resizebox{\linewidth}{!}{%
  \begin{tabular}{@{}lccccccccc@{}}
    \toprule
    & \multicolumn{8}{c}{\textbf{Understanding}} & \textbf{Gen.} \\
    \cmidrule(l){2-10}
    \textbf{Configuration} & \textbf{MMMU} & \textbf{MMB} & \textbf{TVQA} & \textbf{SEED} & \textbf{RWQA} & \textbf{MMVet} & \textbf{MME-P} & \textbf{MME-C} & \textbf{GenEval} \\
    \midrule
    BLIP3o-8B (base) & 50.6 & 83.5 & 83.1 & 77.5 & 69.0 & 66.6 & 1682.6 & 647.1 & 84 \\
    \rowcolor{green!8}
    Full framework (Ours) & \textbf{52.8} & \textbf{86.1} & \textbf{85.2} & \textbf{79.4} & \textbf{70.9} & \textbf{68.7} & \textbf{1698.4} & \textbf{660.3} & \textbf{87} \\
    \midrule
    w/o generation loop (und.\ only) & 52.5 & 85.7 & 84.8 & 79.1 & 70.6 & 68.4 & 1696.0 & 658.3 & 83 \\
    w/o understanding loop (gen.\ only) & 50.6 & 83.5 & 83.1 & 77.5 & 69.0 & 66.6 & 1682.6 & 647.1 & 86 \\
    Self-generated images only & 52.0 & 85.3 & 84.3 & 78.8 & 70.0 & 68.0 & 1694.0 & 656.0 & 85 \\
    \bottomrule
  \end{tabular}%
  }
\end{table}

\noindent\textbf{Parameter Strategy Comparison.} Table~\ref{tab:param_strategy} indicates that adapter updates are more stable than full-parameter updates for our internal rewards. This is a setting-specific result, not a general claim against full fine-tuning.

\begin{table}[!htbp]
  \caption{\textbf{Parameter update strategies on BLIP3o-8B under identical data, rewards, and optimization.} LoRA performs best, QLoRA remains positive but loses some gain, and full fine-tuning underperforms the base (MMMU 50.2, GenEval 83\%), suggesting adapter updates are more stable for these internal rewards.}
  \label{tab:param_strategy}
  \centering
  \setlength{\tabcolsep}{3pt}
  \resizebox{\linewidth}{!}{%
  \begin{tabular}{@{}lcccccccccc@{}}
    \toprule
    & & \multicolumn{8}{c}{\textbf{Understanding}} & \textbf{Generation} \\
    \cmidrule(l){3-11}
    \textbf{Strategy} & \textbf{Params (\%)} & \textbf{MMMU} & \textbf{MMB} & \textbf{TVQA} & \textbf{SEED} & \textbf{RWQA} & \textbf{MMVet} & \textbf{MME-P} & \textbf{MME-C} & \textbf{GenEval} \\
    \midrule
    BLIP3o-8B (base) & -- & 50.6 & 83.5 & 83.1 & 77.5 & 69.0 & 66.6 & 1682.6 & 647.1 & 84 \\
    \midrule
    \rowcolor{green!8}
    \textbf{LoRA} (default) & $\sim$0.8\% & \textbf{52.8} & \textbf{86.1} & \textbf{85.2} & \textbf{79.4} & \textbf{70.9} & \textbf{68.7} & \textbf{1698.4} & \textbf{660.3} & \textbf{87} \\
    QLoRA (4-bit) & $\sim$0.8\% & 52.0 & 85.3 & 84.4 & 78.6 & 70.1 & 67.9 & 1692.8 & 655.7 & 86 \\
    SFT (self-generated data) & 100\% & 51.4 & 84.6 & 83.9 & 78.1 & 69.6 & 67.2 & 1688.3 & 651.4 & 85 \\
    Full fine-tuning & 100\% & 50.2 & 83.1 & 82.8 & 77.0 & 68.5 & 66.0 & 1678.9 & 643.5 & 83 \\
    \bottomrule
  \end{tabular}%
  }
\end{table}

\begin{figure}[H]
  \centering
  \includegraphics[width=0.94\linewidth]{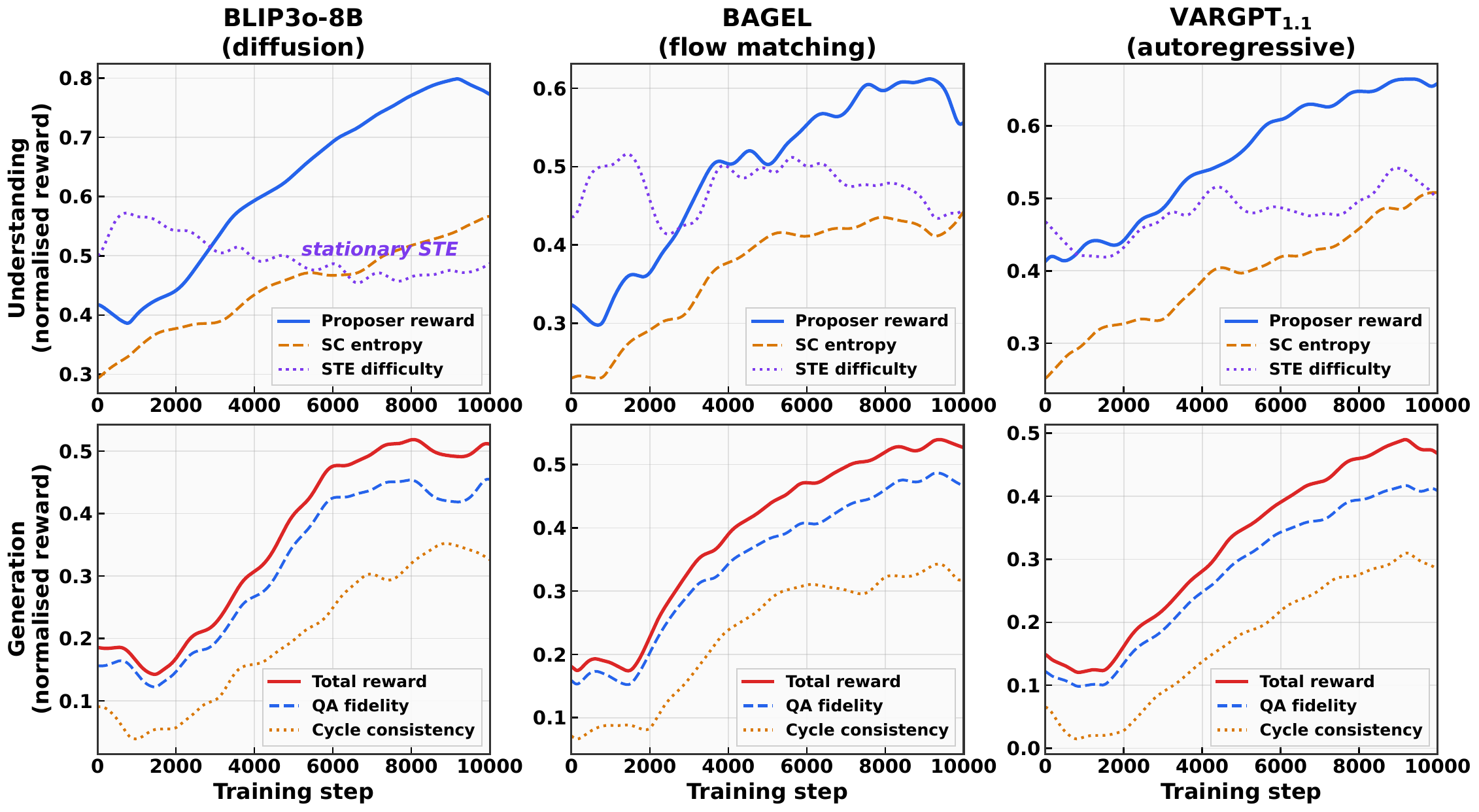}
  \caption{\textbf{Training dynamics over a roughly 10k-step horizon.} Understanding signals stabilize after STE-driven exploration, and generation rewards rise across diffusion, rectified-flow, and autoregressive backbones without reward plateaus.}
  \label{fig:training_dynamics}
\end{figure}

\noindent \textbf{Training Dynamics.} Figure~\ref{fig:training_dynamics} shows STE-driven exploration followed by stable self-consistency and rising generation rewards across all three backbones.

\begin{figure}[H]
  \centering
  \includegraphics[width=\linewidth]{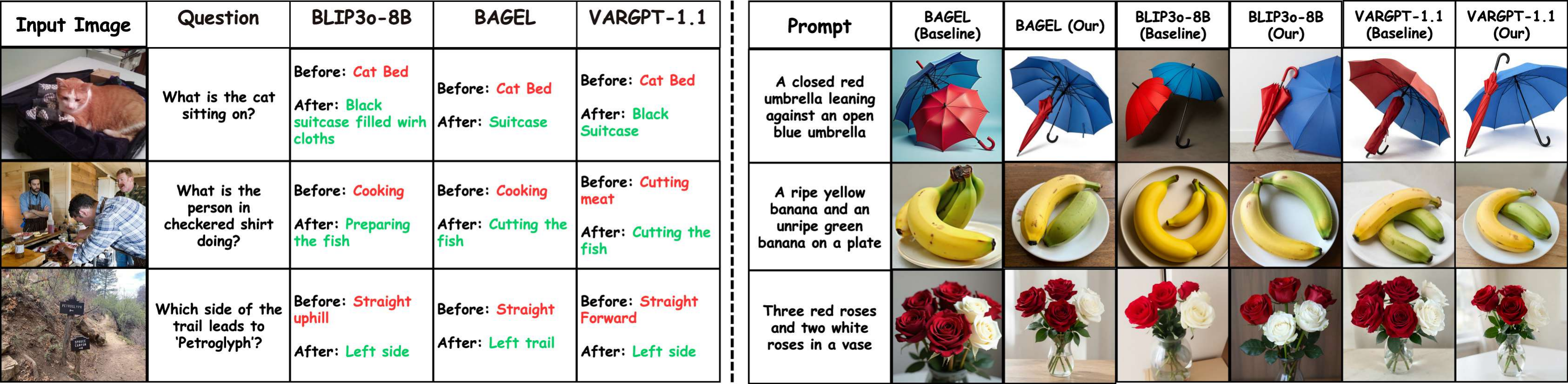}
  \caption{Qualitative comparison of base vs.\ self-evolved outputs across tasks and backbones. \textbf{Left (visual understanding):} Before/after answers for BLIP3o-8B, BAGEL, and VARGPT-v1.1. Incorrect base answers (\textcolor{red}{red}) are corrected after training (\textcolor{ForestGreen}{green}), improving object recognition (cat on suitcase), action understanding (cooking$\to$cutting), and spatial reasoning (uphill$\to$left side). \textbf{Right (image generation):} Baseline vs.\ ours on three compositional prompts. After training, all backbones better satisfy color attributes, object counts, and fine-grained composition (e.g., umbrella color, rose counts), consistent across image generation paradigms.}
  \label{fig:qualitative}
\end{figure}

\noindent \textbf{Qualitative Results.} Figure~\ref{fig:qualitative} shows representative before/after examples: evolved checkpoints correct object, action, and spatial understanding errors and improve color, count, and compositional fidelity in generated images.

\section{Conclusion}
\label{sec:conclusion}

We introduced a self-evolving framework for unified understanding and generation using only unlabeled images and no task-trained external reward/judge model. With Proposer, Solver, and Generator LoRA roles on a frozen backbone, internal consistency yields +1.9 to +3.6 gains on percentage-based understanding metrics, double-digit MME sub-score gains, and +3\% GenEval across BLIP3o-8B, BAGEL, and VARGPT-v1.1 under one matched recipe. Ablations support independent signal contributions and solver-mediated coupling beyond either loop alone. The method is LoRA-based post-training of existing latent capabilities: supervision is bounded by Solver quality, and full fine-tuning is less stable under internal rewards. Future work should study stronger internal evaluators, larger unlabeled pools, and video/3D extensions.


\section{Acknowledgements}
The computations were enabled by resources provided by  LUMI hosted by CSC (Finland) and LUMI consortium, and by Berzelius resource provided by the Knut and Alice Wallenberg Foundation at the NSC.

\bibliographystyle{plainnat}
\bibliography{main}

\clearpage
\appendix
\setcounter{table}{0}
\renewcommand{\thetable}{S\arabic{table}}
\setcounter{figure}{0}
\renewcommand{\thefigure}{S\arabic{figure}}
\setcounter{algorithm}{0}
\renewcommand{\thealgorithm}{S\arabic{algorithm}}

\section*{Appendix}

\begin{table}[!b]
  \caption{Key hyperparameters used in the self-evolving training runs. Shared settings are listed first; backbone-specific entries are marked where applicable.}
  \label{tab:sup_hyperparams}
  \centering
  \small
  \setlength{\tabcolsep}{4pt}
  \renewcommand{\arraystretch}{1.05}
  \begin{tabular}{@{}p{0.17\linewidth}p{0.23\linewidth}p{0.52\linewidth}@{}}
    \toprule
    Category & Hyperparameter & Value \\
    \midrule
    Data & Unlabeled image pool & 10{,}000 images sampled from COCO~\cite{lin2014microsoft}, SA-1B~\cite{kirillov2023segment}, TextVQA~\cite{singh2019textvqa}, GQA~\cite{hudson2019gqa}, and LAION-COCO~\cite{schuhmann2022laioncoco}; all annotations discarded \\
    Optimization & Steps / precision & Roughly 10k steps (10{,}000 for BAGEL/VARGPT; 10{,}250 in the final BLIP3o logs); bfloat16; deterministic; seed 42 \\
    Optimization & Optimizer & AdamW; lr $1\times10^{-6}$; weight decay 0.01; grad clip 1.0; grad accumulation 1 \\
    LoRA & Roles / backbone & 3 role adapters (Proposer, Solver, Generator); frozen backbone \\
    LoRA & Rank / $\alpha$ / dropout & $r=16$, $\alpha=32$, dropout 0.05 \\
    LoRA & Text-role target modules & \texttt{q\_proj,\allowbreak k\_proj,\allowbreak v\_proj,\allowbreak o\_proj,\allowbreak gate\_proj,\allowbreak up\_proj,\allowbreak down\_proj} \\
    LoRA & BLIP3o merger LoRA & \texttt{visual.merger.mlp.0,\allowbreak visual.merger.mlp.2} \\
    LoRA & BLIP3o DiT targets & \texttt{attn2.to\_q,\allowbreak attn2.to\_k,\allowbreak attn2.to\_v,\allowbreak attn2.to\_out.0,\allowbreak caption\_projection.linear\_1,\allowbreak caption\_projection.linear\_2} \\
    Schedule & U:G cycle & 3 visual understanding steps : 2 image generation steps \\
    Understanding & PPS / candidates & $N=7$ prompt perturbations; $K=3$ proposer candidates; 3 spot-check samples \\
    Understanding & STE & maximum next-token entropy over the first 5 answer tokens; rolling window size 128 \\
    Generation (BLIP3o) & Candidates / inference & $L=3$ candidate images; 896$\times$896; 50 inference steps; guidance scale 2.0 \\
    Spec quality & Gates & min spec quality 0.35; min QA pairs 2 \\
    Rewards & Weights / penalty & QA fidelity 0.65; cycle consistency 0.20; diversity 0.10; contradiction penalty 0.20 \\
    Token-policy KL & Coef / target / bounds & coef 0.01; target 0.02; adapt rate 0.10; bounds $[10^{-3},10^{2}]$ \\
    \bottomrule
  \end{tabular}
\end{table}

\noindent We provide additional details needed to reproduce and interpret the experiments. The appendix includes the training algorithm, implementation settings, reward definitions, hyperparameter sweeps, control experiments, diagnostic analyses, prompt templates, and additional qualitative examples.

\section{Training Algorithm}
\label{sec:sup_algorithm}

Algorithm~\ref{alg:self_evolving} summarizes the end-to-end training loop. We keep the backbone frozen and update only three LoRA adapters (Proposer, Solver, Generator). Proposer/Solver updates use KL-regularized policy gradients, while Generator updates use the same internal reward but are routed through each backbone's native generator objective: token-policy updates for autoregressive generators and reward-weighted denoising for diffusion/flow generators. The schedule alternates between visual understanding and image generation steps (default 3:2).

\begin{algorithm}[!htb]
  \caption{Self-evolving UUG training with coupled visual understanding and image generation loops. The algorithm alternates between question proposal/answering and reward-driven generation updates while keeping the backbone frozen and training only the three LoRA role adapters.}
  \label{alg:self_evolving}
  \small
  \begin{algorithmic}[1]
    \STATE \textbf{Input:} unlabeled image pool $\mathcal{D}$; frozen backbone $\mathcal{M}_\theta$; LoRA adapters $\phi_p,\phi_s,\phi_g$
    \STATE \textbf{Initialize:} reference policies $\pi_{\theta}$; STE window $\mathcal{W}$; (optional) replay buffer $\mathcal{B}$
    \FOR{$\text{step}=1$ \TO $T$}
      \IF{visual understanding phase}
        \STATE Sample image $\mathcal{I}\sim\mathcal{D}$ (optionally mix $\mathcal{I}\sim\mathcal{B}$)
        \STATE Proposer samples $K$ candidate questions $\{q_k\}_{k=1}^K \sim \pi_{\phi_p}(\cdot\mid\mathcal{I})$
        \FOR{$k=1$ \TO $K$}
          \STATE Solver answers under $N$ prompt perturbations: $a_{k,i}\sim\pi_{\phi_s}(\cdot\mid\mathcal{I},\rho_i(q_k))$
          \STATE Compute self-consistency entropy $H_{\text{sc}}(q_k)$ from $\{a_{k,i}\}_{i=1}^N$
          \STATE Compute STE difficulty $d_{\text{ste}}(q_k)$ from a greedy solver call; normalize via rolling-window quantile in $\mathcal{W}$
          \STATE Score candidate with frontier-seeking reward $R_p(q_k)$ (blend of self-consistency and STE, plus light penalties)
        \ENDFOR
        \STATE Update Proposer $\phi_p$ with GRPO over the $K$ candidates using $\{R_p(q_k)\}_{k=1}^K$ \cite{shao2024deepseekmath}
        \STATE Choose $q^\star=\arg\max_k R_p(q_k)$ and update Solver $\phi_s$ with REINFORCE using agreement/format reward $R_s(\mathcal{I},q^\star)$
      \ELSE
        \STATE \textit{Image generation phase}
        \STATE Sample real image $\mathcal{I}\sim\mathcal{D}$ and produce prompt $t$ (caption $\mathcal{I}$)
        \STATE Sample $M$ diagnostic questions $\{q_j\}_{j=1}^M$ and Solver-derived reference answers $\{a_j^{\text{ref}}\}$ on $\mathcal{I}$
        \STATE Generator samples $L$ candidate images $\{\hat{\mathcal{I}}_\ell\}_{\ell=1}^L \sim \pi_{\phi_g}(\cdot\mid t)$
        \STATE Score each candidate via QA fidelity + cycle consistency (+ diversity/contradiction) to obtain rewards $\{R_g(\hat{\mathcal{I}}_\ell)\}$
        \STATE Update Generator $\phi_g$ with the native generator objective using $\{R_g(\hat{\mathcal{I}}_\ell)\}$
        \STATE Push the best candidate into replay buffer $\mathcal{B}$ if it passes the quality gate
      \ENDIF
    \ENDFOR
  \end{algorithmic}
\end{algorithm}

\section{Experimental Details}
\label{sec:sup_implementation}

\noindent\textbf{Core optimization settings.}
Reported runs use a roughly 10k-step training horizon with AdamW in bfloat16, learning rate $1\times10^{-6}$, weight decay 0.01, gradient clip 1.0, and gradient accumulation 1. The BAGEL and VARGPT runs use 10{,}000 steps, while the final BLIP3o logs extend to 10{,}250 steps. We freeze the backbone and train role LoRA adapters~\cite{hu2022lora} with rank 16 ($\alpha=32$, dropout 0.05). Text-role adapters target q\_proj, k\_proj, v\_proj, o\_proj, gate\_proj, up\_proj, and down\_proj; BLIP3o additionally uses generator-side DiT LoRA targets listed in Table~\ref{tab:sup_hyperparams}.

\noindent\textbf{Sampling and generation settings.}
Understanding uses $N=7$ prompt perturbations (Prompt-Perturbed Sampling) and $K=3$ proposer candidates per step, with 3 spot-check samples for candidate selection. For generation, we use $L=3$ candidate images per prompt, image size 896$\times$896, 50 denoising/inference steps, and guidance scale 2.0 for BLIP3o-based runs~\cite{chen2025blip3o}.

\noindent\textbf{Reward and regularization.}
Generation reward uses QA fidelity weight 0.65, cycle-consistency weight 0.20, diversity weight 0.10, and a contradiction penalty weight of 0.20. Quality gates use minimum spec quality 0.35 and minimum 2 QA pairs per sample. Token-policy KL control uses coefficient 0.01, target 0.02, adaptation rate 0.10, and bounds $[10^{-3},10^2]$.

\noindent\textbf{Cycle schedules.}
The default joint run uses a 3:2 visual understanding\,:\,image generation cycle. Visual understanding-only and image generation-only ablations use 5:0 and 0:5, respectively. The self-generated-images variant also uses 3:2, with replay-buffer-based training enabled (buffer size 500; generated mix ratio fixed to 1.0).

\noindent\textbf{Determinism and logging.}
All final scripts run with deterministic mode enabled, save checkpoints every 50 steps, and use seed 42 in the default configuration.

\section{Operational Definitions of Reward Terms}
\label{sec:sup_reward_terms}

This section clarifies how the reward terms named in the main method are instantiated in the training code.

\noindent\textbf{Understanding-side self-consistency and majority fraction.}
For each question, the Solver produces $N$ answers under prompt-perturbed sampling. Answers are normalized before voting by lowercasing, stripping punctuation, collapsing whitespace, and truncating to a short core span so phrasing differences do not create artificial disagreement. The majority fraction is then the vote share of the most frequent normalized answer, and self-consistency entropy is the Shannon entropy of the empirical answer distribution over these normalized votes.

\noindent\textbf{Solver Token Entropy (STE).}
STE is computed from the Solver's token probabilities rather than from answer voting. Concretely, we measure the full softmax entropy at each generated answer token and use the maximum value over the first five answer tokens as the raw uncertainty signal. This raw value is converted into a difficulty score through a rolling-window rank statistic, so a question receives a high STE score when it triggers higher token-level uncertainty than most recent questions, even if the final sampled answers collapse to the same surface form.

\noindent\textbf{Generation-side QA fidelity and contradiction.}
Each proposed generation spec contains verification QA pairs. For a generated image, the Solver answers each verification question multiple times, we take the majority answer, and compare it to the expected answer with a soft-match rule: exact match scores 1.0, substring containment scores 0.8, near-numeric matches receive partial credit, and otherwise lexical overlap is used. The per-question QA fidelity score is a weighted combination of answer match (0.7) and majority fraction (0.3). Contradiction is a separate penalty that activates for explicit yes/no polarity conflicts between the predicted and expected answers.

\noindent\textbf{Cycle consistency, diversity, and final generator reward.}
Cycle consistency captions the generated image and measures semantic agreement between the caption, the prompt, and the image using fixed embedding-based similarities (model-native embeddings where available and CLIP-style image-text similarity only in wrappers that require it), with token-overlap fallback only if embedding computation fails. Diversity is computed across candidate images in the same generation batch using leave-one-out image diversity, so candidates are rewarded only when they add non-redundant variation. The final generation reward is the weighted sum of QA fidelity, cycle consistency, and diversity, minus contradiction, and is then multiplied by the spec-quality gate so low-quality prompt cards cannot receive high reward solely from image-side scoring.

\section{Hyperparameter Details}
\label{sec:sup_hyperparams}

Table~\ref{tab:sup_hyperparams} lists the adopted training defaults used across the main experiments. Tables~\ref{tab:sup_hparam_understanding}, \ref{tab:sup_hparam_optimization}, and \ref{tab:sup_hparam_generation} then report one-factor sweeps for understanding-side design choices, optimization settings, and generation-side reward settings, respectively. Together, these tables show both the exact configuration used in the reported runs and the local sensitivity of the method around that operating point. We use these one-factor sweeps to select an adopted configuration and do not perform exhaustive combinational hyperparameter search.

\begin{table*}[!htb]
  \caption{Understanding-side hyperparameter sweeps on BLIP3o-8B under the shared roughly 10k-step protocol. Each block varies one understanding-side factor at a time. Deltas are relative to the BLIP3o-8B base checkpoint. Shaded rows marked with $\dagger$ denote the adopted setting; all other hyperparameters follow Table~\ref{tab:sup_hyperparams}.}
  \label{tab:sup_hparam_understanding}
  \centering
  \scriptsize
  \setlength{\tabcolsep}{3pt}
  \resizebox{\linewidth}{!}{%
  \begin{tabular}{@{}llccccccccc@{}}
    \toprule
    & & \multicolumn{8}{c}{\textbf{Visual understanding}} & \textbf{Image gen.} \\
    \cmidrule(l){3-11}
    \textbf{Variant} & \textbf{Setting} & \textbf{MMMU} & \textbf{MMB} & \textbf{TVQA} & \textbf{SEED} & \textbf{RWQA} & \textbf{MMVet} & \textbf{MME-P} & \textbf{MME-C} & \textbf{GenEval overall (\%)} \\
    \midrule
    BLIP3o-8B (base) & --- & 50.6 & 83.5 & 83.1 & 77.5 & 69.0 & 66.6 & 1682.6 & 647.1 & 84\% \\
    \midrule
    \noalign{\vskip 1.5pt}
    \groupcell{5}{LoRA rank $r$} & 4 & 51.8 \dpos{1.2} & 85.4 \dpos{1.9} & 84.9 \dpos{1.8} & 78.6 \dpos{1.1} & 70.5 \dpos{1.5} & 67.8 \dpos{1.2} & 1692.6 \dpos{10.0} & 654.0 \dpos{6.9} & 85\% \dpos{1\%} \\
    & 8 & 52.4 \dpos{1.8} & 85.9 \dpos{2.4} & 85.1 \dpos{2.0} & 79.5 \dpos{2.0} & 70.7 \dpos{1.7} & 68.4 \dpos{1.8} & 1696.3 \dpos{13.7} & 657.5 \dpos{10.4} & 86\% \dpos{2\%} \\
    & \adopted{$16^{\dagger}$} & \adopted{52.8 \dpos{2.2}} & \adopted{86.1 \dpos{2.6}} & \adopted{85.2 \dpos{2.1}} & \adopted{79.4 \dpos{1.9}} & \adopted{70.9 \dpos{1.9}} & \adopted{68.7 \dpos{2.1}} & \adopted{1698.4 \dpos{15.8}} & \adopted{660.3 \dpos{13.2}} & \adopted{87\% \dpos{3\%}} \\
    & 32 & 52.6 \dpos{2.0} & 86.2 \dpos{2.7} & 85.2 \dpos{2.1} & 79.3 \dpos{1.8} & 70.8 \dpos{1.8} & 68.7 \dpos{2.1} & 1697.6 \dpos{15.0} & 660.6 \dpos{13.5} & 87\% \dpos{3\%} \\
    & 64 & 52.2 \dpos{1.6} & 85.8 \dpos{2.3} & 85.0 \dpos{1.9} & 79.0 \dpos{1.5} & 70.6 \dpos{1.6} & 68.8 \dpos{2.2} & 1695.7 \dpos{13.1} & 656.8 \dpos{9.7} & 86\% \dpos{2\%} \\
    \midrule
    \noalign{\vskip 1.5pt}
    \groupcell{5}{PPS count $N$} & 3 & 52.1 \dpos{1.5} & 85.6 \dpos{2.1} & 85.3 \dpos{2.2} & 79.0 \dpos{1.5} & 70.6 \dpos{1.6} & 68.1 \dpos{1.5} & 1694.8 \dpos{12.2} & 656.2 \dpos{9.1} & 86\% \dpos{2\%} \\
    & 5 & 52.5 \dpos{1.9} & 86.0 \dpos{2.5} & 85.1 \dpos{2.0} & 79.2 \dpos{1.7} & 70.9 \dpos{1.9} & 68.5 \dpos{1.9} & 1697.0 \dpos{14.4} & 658.4 \dpos{11.3} & 87\% \dpos{3\%} \\
    & \adopted{$7^{\dagger}$} & \adopted{52.8 \dpos{2.2}} & \adopted{86.1 \dpos{2.6}} & \adopted{85.2 \dpos{2.1}} & \adopted{79.4 \dpos{1.9}} & \adopted{70.9 \dpos{1.9}} & \adopted{68.7 \dpos{2.1}} & \adopted{1698.4 \dpos{15.8}} & \adopted{660.3 \dpos{13.2}} & \adopted{87\% \dpos{3\%}} \\
    & 9 & 53.0 \dpos{2.4} & 86.1 \dpos{2.6} & 85.2 \dpos{2.1} & 79.3 \dpos{1.8} & 70.8 \dpos{1.8} & 68.5 \dpos{1.9} & 1698.9 \dpos{16.3} & 660.0 \dpos{12.9} & 87\% \dpos{3\%} \\
    & 11 & 52.9 \dpos{2.3} & 85.9 \dpos{2.4} & 85.2 \dpos{2.1} & 79.3 \dpos{1.8} & 70.7 \dpos{1.7} & 68.4 \dpos{1.8} & 1698.2 \dpos{15.6} & 659.5 \dpos{12.4} & 86\% \dpos{2\%} \\
    \midrule
    \noalign{\vskip 1.5pt}
    \groupcell{3}{Proposer cand. $K$} & 1 & 51.9 \dpos{1.3} & 85.5 \dpos{2.0} & 85.0 \dpos{1.9} & 78.7 \dpos{1.2} & 70.4 \dpos{1.4} & 68.0 \dpos{1.4} & 1693.9 \dpos{11.3} & 655.1 \dpos{8.0} & 85\% \dpos{1\%} \\
    & \adopted{$3^{\dagger}$} & \adopted{52.8 \dpos{2.2}} & \adopted{86.1 \dpos{2.6}} & \adopted{85.2 \dpos{2.1}} & \adopted{79.4 \dpos{1.9}} & \adopted{70.9 \dpos{1.9}} & \adopted{68.7 \dpos{2.1}} & \adopted{1698.4 \dpos{15.8}} & \adopted{660.3 \dpos{13.2}} & \adopted{87\% \dpos{3\%}} \\
    & 5 & 53.0 \dpos{2.4} & 86.1 \dpos{2.6} & 85.1 \dpos{2.0} & 79.5 \dpos{2.0} & 70.9 \dpos{1.9} & 68.6 \dpos{2.0} & 1697.9 \dpos{15.3} & 660.7 \dpos{13.6} & 87\% \dpos{3\%} \\
    \midrule
    \noalign{\vskip 1.5pt}
    \groupcell{3}{STE window $W$} & 64 & 52.3 \dpos{1.7} & 85.9 \dpos{2.4} & 85.1 \dpos{2.0} & 79.1 \dpos{1.6} & 70.6 \dpos{1.6} & 68.3 \dpos{1.7} & 1696.8 \dpos{14.2} & 657.9 \dpos{10.8} & 86\% \dpos{2\%} \\
    & \adopted{$128^{\dagger}$} & \adopted{52.8 \dpos{2.2}} & \adopted{86.1 \dpos{2.6}} & \adopted{85.2 \dpos{2.1}} & \adopted{79.4 \dpos{1.9}} & \adopted{70.9 \dpos{1.9}} & \adopted{68.7 \dpos{2.1}} & \adopted{1698.4 \dpos{15.8}} & \adopted{660.3 \dpos{13.2}} & \adopted{87\% \dpos{3\%}} \\
    & 256 & 52.7 \dpos{2.1} & 86.1 \dpos{2.6} & 85.3 \dpos{2.2} & 79.5 \dpos{2.0} & 70.9 \dpos{1.9} & 68.6 \dpos{2.0} & 1698.7 \dpos{16.1} & 659.7 \dpos{12.6} & 87\% \dpos{3\%} \\
    \bottomrule
  \end{tabular}%
  }
\end{table*}

\begin{table*}[!htb]
  \caption{Optimization and regularization sweeps on BLIP3o-8B under the shared roughly 10k-step protocol. Each block varies one optimization factor at a time. Deltas are relative to the BLIP3o-8B base checkpoint. Shaded rows marked with $\dagger$ denote the adopted setting; all other hyperparameters follow Table~\ref{tab:sup_hyperparams}.}
  \label{tab:sup_hparam_optimization}
  \centering
  \scriptsize
  \setlength{\tabcolsep}{3pt}
  \resizebox{\linewidth}{!}{%
  \begin{tabular}{@{}llccccccccc@{}}
    \toprule
    & & \multicolumn{8}{c}{\textbf{Visual understanding}} & \textbf{Image gen.} \\
    \cmidrule(l){3-11}
    \textbf{Variant} & \textbf{Setting} & \textbf{MMMU} & \textbf{MMB} & \textbf{TVQA} & \textbf{SEED} & \textbf{RWQA} & \textbf{MMVet} & \textbf{MME-P} & \textbf{MME-C} & \textbf{GenEval overall (\%)} \\
    \midrule
    BLIP3o-8B (base) & --- & 50.6 & 83.5 & 83.1 & 77.5 & 69.0 & 66.6 & 1682.6 & 647.1 & 84\% \\
    \midrule
    \noalign{\vskip 1.5pt}
    \groupcell{3}{Learning rate} & $5\times10^{-7}$ & 52.2 \dpos{1.6} & 85.7 \dpos{2.2} & 85.0 \dpos{1.9} & 79.1 \dpos{1.6} & 70.6 \dpos{1.6} & 68.2 \dpos{1.6} & 1695.6 \dpos{13.0} & 656.8 \dpos{9.7} & 86\% \dpos{2\%} \\
    & \adopted{${1\times10^{-6}}^{\dagger}$} & \adopted{52.8 \dpos{2.2}} & \adopted{86.1 \dpos{2.6}} & \adopted{85.2 \dpos{2.1}} & \adopted{79.4 \dpos{1.9}} & \adopted{70.9 \dpos{1.9}} & \adopted{68.7 \dpos{2.1}} & \adopted{1698.4 \dpos{15.8}} & \adopted{660.3 \dpos{13.2}} & \adopted{87\% \dpos{3\%}} \\
    & $2\times10^{-6}$ & 52.5 \dpos{1.9} & 86.0 \dpos{2.5} & 84.8 \dpos{1.7} & 79.2 \dpos{1.7} & 71.0 \dpos{2.0} & 68.3 \dpos{1.7} & 1696.2 \dpos{13.6} & 657.2 \dpos{10.1} & 85\% \dpos{1\%} \\
    \midrule
    \noalign{\vskip 1.5pt}
    \groupcell{3}{Weight decay} & 0.00 & 52.6 \dpos{2.0} & 86.0 \dpos{2.5} & 85.3 \dpos{2.2} & 79.3 \dpos{1.8} & 70.8 \dpos{1.8} & 68.4 \dpos{1.8} & 1697.4 \dpos{14.8} & 658.5 \dpos{11.4} & 86\% \dpos{2\%} \\
    & \adopted{$0.01^{\dagger}$} & \adopted{52.8 \dpos{2.2}} & \adopted{86.1 \dpos{2.6}} & \adopted{85.2 \dpos{2.1}} & \adopted{79.4 \dpos{1.9}} & \adopted{70.9 \dpos{1.9}} & \adopted{68.7 \dpos{2.1}} & \adopted{1698.4 \dpos{15.8}} & \adopted{660.3 \dpos{13.2}} & \adopted{87\% \dpos{3\%}} \\
    & 0.05 & 52.5 \dpos{1.9} & 85.9 \dpos{2.4} & 85.1 \dpos{2.0} & 79.2 \dpos{1.7} & 70.9 \dpos{1.9} & 68.5 \dpos{1.9} & 1698.6 \dpos{16.0} & 658.4 \dpos{11.3} & 86\% \dpos{2\%} \\
    \midrule
    \noalign{\vskip 1.5pt}
    \groupcell{3}{LoRA dropout} & 0.00 & 52.7 \dpos{2.1} & 86.1 \dpos{2.6} & 85.3 \dpos{2.2} & 79.4 \dpos{1.9} & 70.8 \dpos{1.8} & 68.6 \dpos{2.0} & 1697.9 \dpos{15.3} & 659.1 \dpos{12.0} & 87\% \dpos{3\%} \\
    & \adopted{$0.05^{\dagger}$} & \adopted{52.8 \dpos{2.2}} & \adopted{86.1 \dpos{2.6}} & \adopted{85.2 \dpos{2.1}} & \adopted{79.4 \dpos{1.9}} & \adopted{70.9 \dpos{1.9}} & \adopted{68.7 \dpos{2.1}} & \adopted{1698.4 \dpos{15.8}} & \adopted{660.3 \dpos{13.2}} & \adopted{87\% \dpos{3\%}} \\
    & 0.10 & 52.6 \dpos{2.0} & 86.1 \dpos{2.6} & 85.1 \dpos{2.0} & 79.2 \dpos{1.7} & 70.7 \dpos{1.7} & 68.8 \dpos{2.2} & 1697.3 \dpos{14.7} & 659.7 \dpos{12.6} & 86\% \dpos{2\%} \\
    \midrule
    \noalign{\vskip 1.5pt}
    \groupcell{3}{KL coeff.} & 0.005 & 52.4 \dpos{1.8} & 85.8 \dpos{2.3} & 85.1 \dpos{2.0} & 79.1 \dpos{1.6} & 70.7 \dpos{1.7} & 68.2 \dpos{1.6} & 1696.4 \dpos{13.8} & 657.5 \dpos{10.4} & 85\% \dpos{1\%} \\
    & \adopted{$0.01^{\dagger}$} & \adopted{52.8 \dpos{2.2}} & \adopted{86.1 \dpos{2.6}} & \adopted{85.2 \dpos{2.1}} & \adopted{79.4 \dpos{1.9}} & \adopted{70.9 \dpos{1.9}} & \adopted{68.7 \dpos{2.1}} & \adopted{1698.4 \dpos{15.8}} & \adopted{660.3 \dpos{13.2}} & \adopted{87\% \dpos{3\%}} \\
    & 0.020 & 52.5 \dpos{1.9} & 86.0 \dpos{2.5} & 85.0 \dpos{1.9} & 79.5 \dpos{2.0} & 70.8 \dpos{1.8} & 68.5 \dpos{1.9} & 1697.0 \dpos{14.4} & 659.4 \dpos{12.3} & 86\% \dpos{2\%} \\
    \bottomrule
  \end{tabular}%
  }
\end{table*}

\begin{table*}[!htb]
  \caption{Generation-side and reward-design sweeps on BLIP3o-8B under the shared roughly 10k-step protocol. Each block varies one generation-side factor at a time. Deltas are relative to the BLIP3o-8B base checkpoint. Shaded rows marked with $\dagger$ denote the adopted setting; all other hyperparameters follow Table~\ref{tab:sup_hyperparams}.}
  \label{tab:sup_hparam_generation}
  \centering
  \scriptsize
  \setlength{\tabcolsep}{3pt}
  \resizebox{\linewidth}{!}{%
  \begin{tabular}{@{}llccccccccc@{}}
    \toprule
    & & \multicolumn{8}{c}{\textbf{Visual understanding}} & \textbf{Image gen.} \\
    \cmidrule(l){3-11}
    \textbf{Variant} & \textbf{Setting} & \textbf{MMMU} & \textbf{MMB} & \textbf{TVQA} & \textbf{SEED} & \textbf{RWQA} & \textbf{MMVet} & \textbf{MME-P} & \textbf{MME-C} & \textbf{GenEval overall (\%)} \\
    \midrule
    BLIP3o-8B (base) & --- & 50.6 & 83.5 & 83.1 & 77.5 & 69.0 & 66.6 & 1682.6 & 647.1 & 84\% \\
    \midrule
    \noalign{\vskip 1.5pt}
    \groupcell{3}{Generations $L$} & 1 & 52.8 \dpos{2.2} & 86.1 \dpos{2.6} & 85.1 \dpos{2.0} & 79.5 \dpos{2.0} & 70.9 \dpos{1.9} & 68.6 \dpos{2.0} & 1698.1 \dpos{15.5} & 660.5 \dpos{13.4} & 84\% \dpos{0\%} \\
    & \adopted{$3^{\dagger}$} & \adopted{52.8 \dpos{2.2}} & \adopted{86.1 \dpos{2.6}} & \adopted{85.2 \dpos{2.1}} & \adopted{79.4 \dpos{1.9}} & \adopted{70.9 \dpos{1.9}} & \adopted{68.7 \dpos{2.1}} & \adopted{1698.4 \dpos{15.8}} & \adopted{660.3 \dpos{13.2}} & \adopted{87\% \dpos{3\%}} \\
    & 5 & 52.9 \dpos{2.3} & 86.1 \dpos{2.6} & 85.2 \dpos{2.1} & 79.4 \dpos{1.9} & 70.8 \dpos{1.8} & 68.7 \dpos{2.1} & 1698.8 \dpos{16.2} & 660.1 \dpos{13.0} & 87\% \dpos{3\%} \\
    \midrule
    \noalign{\vskip 1.5pt}
    \groupcell{3}{Min.\ QA pairs} & 1 & 52.8 \dpos{2.2} & 86.2 \dpos{2.7} & 85.2 \dpos{2.1} & 79.3 \dpos{1.8} & 70.9 \dpos{1.9} & 68.6 \dpos{2.0} & 1698.2 \dpos{15.6} & 660.6 \dpos{13.5} & 85\% \dpos{1\%} \\
    & \adopted{$2^{\dagger}$} & \adopted{52.8 \dpos{2.2}} & \adopted{86.1 \dpos{2.6}} & \adopted{85.2 \dpos{2.1}} & \adopted{79.4 \dpos{1.9}} & \adopted{70.9 \dpos{1.9}} & \adopted{68.7 \dpos{2.1}} & \adopted{1698.4 \dpos{15.8}} & \adopted{660.3 \dpos{13.2}} & \adopted{87\% \dpos{3\%}} \\
    & 3 & 52.7 \dpos{2.1} & 86.1 \dpos{2.6} & 85.3 \dpos{2.2} & 79.4 \dpos{1.9} & 70.9 \dpos{1.9} & 68.8 \dpos{2.2} & 1698.7 \dpos{16.1} & 659.9 \dpos{12.8} & 86\% \dpos{2\%} \\
    \midrule
    \noalign{\vskip 1.5pt}
    \groupcell{3}{\parbox{1.35cm}{\centering QA fidelity\\$w_{\text{fid}}$}} & 0.50 & 52.9 \dpos{2.3} & 86.1 \dpos{2.6} & 85.1 \dpos{2.0} & 79.4 \dpos{1.9} & 71.0 \dpos{2.0} & 68.7 \dpos{2.1} & 1698.3 \dpos{15.7} & 659.8 \dpos{12.7} & 85\% \dpos{1\%} \\
    & \adopted{$0.65^{\dagger}$} & \adopted{52.8 \dpos{2.2}} & \adopted{86.1 \dpos{2.6}} & \adopted{85.2 \dpos{2.1}} & \adopted{79.4 \dpos{1.9}} & \adopted{70.9 \dpos{1.9}} & \adopted{68.7 \dpos{2.1}} & \adopted{1698.4 \dpos{15.8}} & \adopted{660.3 \dpos{13.2}} & \adopted{87\% \dpos{3\%}} \\
    & 0.80 & 52.7 \dpos{2.1} & 86.1 \dpos{2.6} & 85.3 \dpos{2.2} & 79.4 \dpos{1.9} & 70.9 \dpos{1.9} & 68.5 \dpos{1.9} & 1698.0 \dpos{15.4} & 660.5 \dpos{13.4} & 86\% \dpos{2\%} \\
    \midrule
    \noalign{\vskip 1.5pt}
    \groupcell{3}{\parbox{1.2cm}{\centering Cycle weight\\$w_{\text{cyc}}$}} & 0.10 & 52.8 \dpos{2.2} & 86.2 \dpos{2.7} & 85.2 \dpos{2.1} & 79.3 \dpos{1.8} & 70.9 \dpos{1.9} & 68.7 \dpos{2.1} & 1698.6 \dpos{16.0} & 660.0 \dpos{12.9} & 85\% \dpos{1\%} \\
    & \adopted{$0.20^{\dagger}$} & \adopted{52.8 \dpos{2.2}} & \adopted{86.1 \dpos{2.6}} & \adopted{85.2 \dpos{2.1}} & \adopted{79.4 \dpos{1.9}} & \adopted{70.9 \dpos{1.9}} & \adopted{68.7 \dpos{2.1}} & \adopted{1698.4 \dpos{15.8}} & \adopted{660.3 \dpos{13.2}} & \adopted{87\% \dpos{3\%}} \\
    & 0.30 & 52.7 \dpos{2.1} & 86.1 \dpos{2.6} & 85.2 \dpos{2.1} & 79.4 \dpos{1.9} & 71.0 \dpos{2.0} & 68.6 \dpos{2.0} & 1697.8 \dpos{15.2} & 660.4 \dpos{13.3} & 86\% \dpos{2\%} \\
    \midrule
    \noalign{\vskip 1.5pt}
    \groupcell{3}{Min.\ spec quality} & 0.25 & 52.8 \dpos{2.2} & 86.0 \dpos{2.5} & 85.2 \dpos{2.1} & 79.5 \dpos{2.0} & 70.9 \dpos{1.9} & 68.6 \dpos{2.0} & 1697.9 \dpos{15.3} & 660.5 \dpos{13.4} & 85\% \dpos{1\%} \\
    & \adopted{$0.35^{\dagger}$} & \adopted{52.8 \dpos{2.2}} & \adopted{86.1 \dpos{2.6}} & \adopted{85.2 \dpos{2.1}} & \adopted{79.4 \dpos{1.9}} & \adopted{70.9 \dpos{1.9}} & \adopted{68.7 \dpos{2.1}} & \adopted{1698.4 \dpos{15.8}} & \adopted{660.3 \dpos{13.2}} & \adopted{87\% \dpos{3\%}} \\
    & 0.45 & 52.8 \dpos{2.2} & 86.1 \dpos{2.6} & 85.3 \dpos{2.2} & 79.3 \dpos{1.8} & 70.9 \dpos{1.9} & 68.7 \dpos{2.1} & 1698.7 \dpos{16.1} & 659.8 \dpos{12.7} & 86\% \dpos{2\%} \\
    \bottomrule
  \end{tabular}%
  }
\end{table*}

\section{Additional Controls and Clarifications}
\label{sec:sup_release_clarifications}

\noindent\textbf{Solver-mediated coupling.}
The coupling between visual understanding and image generation should be read as \emph{solver-mediated} rather than as symmetric gradient exchange between all roles. Proposer/Solver updates improve the Solver's answers and uncertainty estimates, and the same evolving Solver supplies generation-side QA-fidelity and cycle-consistency rewards. This makes the direct mechanism strongest from understanding to generation, while the joint schedule still trains the three role adapters under one alternating procedure. Table~\ref{tab:sup_coupling_controls} adds controls including single-loop variants, training on self-generated images, and a two-stage alternative that trains Proposer/Solver before Generator updates. Joint training gives the best understanding/generation trade-off on both BLIP3o-8B and BAGEL, but we interpret this as evidence for solver-mediated coupling rather than a claim of direct generation-to-understanding gradient feedback.

\begin{table*}[!htb]
  \caption{Additional coupling controls. Understanding average is computed over MMMU, MMBench, TextVQA, SEED-Bench, RealWorldQA, and MM-Vet; GenEval is the overall compositional generation score. The controls use the same unlabeled image pool and compare single-loop training, self-generated-image training, a two-stage Proposer/Solver then Generator schedule, and the alternating joint schedule.}
  \label{tab:sup_coupling_controls}
  \centering
  \small
  \setlength{\tabcolsep}{5pt}
  \resizebox{\linewidth}{!}{%
  \begin{tabular}{@{}lccccc@{}}
    \toprule
    \multirow{2}{*}{Setting} & \multirow{2}{*}{Learnable roles} & \multicolumn{2}{c}{BLIP3o-8B} & \multicolumn{2}{c}{BAGEL} \\
    \cmidrule(lr){3-4}\cmidrule(l){5-6}
     & & Und. Avg. & GenEval & Und. Avg. & GenEval \\
    \midrule
    Base & -- & 71.7 & 84 & 74.0 & 82 \\
    Understanding-only & Proposer/Solver & 73.5 \dpos{1.8} & 83 \dneg{1} & 75.7 \dpos{1.7} & 81 \dneg{1} \\
    Generation-only & Generator & 71.7 \dpos{0.0} & 86 \dpos{2} & 74.0 \dpos{0.0} & 83 \dpos{1} \\
    Self-generated pool & All roles & 73.1 \dpos{1.4} & 85 \dpos{1} & 75.4 \dpos{1.4} & 83 \dpos{1} \\
    Two-stage & Proposer/Solver then Generator & 72.9 \dpos{1.2} & 85 \dpos{1} & 75.7 \dpos{1.7} & 83 \dpos{1} \\
    \rowcolor{green!8}
    \textbf{Joint (ours)} & \textbf{All roles} & \textbf{73.9 \dpos{2.2}} & \textbf{87 \dpos{3}} & \textbf{76.6 \dpos{2.6}} & \textbf{85 \dpos{3}} \\
    \bottomrule
  \end{tabular}%
  }
\end{table*}

\noindent\textbf{STE aggregation and prompt framing.}
The default Solver Token Entropy signal is the maximum full-softmax entropy over the first five generated answer tokens. This max aggregation is intended to detect a decisive uncertain token in short answers, such as count, color, relation, or negation tokens. In the final BLIP3o roughly 10k rollout logs, the normalized Solver answers are short (43{,}197 answers; median 1 word, 90th percentile 3 words, 99th percentile 7 words; 97.6\% are at most 5 words and 100.0\% are at most 10 words), so the entropy window covers the answer-local span in nearly all cases. The implementation also exposes mean aggregation as a control, but all reported main runs use max aggregation. Prompt-perturbed sampling changes only the instruction preamble; the free-form question is inserted verbatim across framings. Therefore, self-consistency uses normalized answer votes across semantically equivalent framings, while STE is a scalar token-uncertainty signal from the Solver and does not require cross-answer token alignment.

\noindent\textbf{Reference answers and reward filtering.}
Generation-side QA references are Solver-derived answers on the real/source image, not labels proposed by the Proposer and not dataset annotations. Low-quality generation specs are filtered before reward computation through the spec-quality gate and minimum-QA-pair requirement. This reduces the chance that a generated image receives a high reward by matching an underspecified prompt card. The QA-fidelity term then checks local attributes, counts, and relations; cycle consistency checks global scene semantics; diversity discourages duplicate candidates; and contradiction penalizes explicit prompt--caption or question--answer polarity conflicts.

\noindent\textbf{Data overlap, scope, and compute.}
For datasets such as TextVQA and GQA, the training pool uses only raw images; labels, boxes, captions, and question-answer annotations are discarded. Any overlap with evaluation sources is therefore distributional rather than supervised. The method should also be interpreted as LoRA-based post-training that better aligns existing latent capabilities of a frozen unified model; it is not a claim that a frozen backbone can learn unlimited new visual primitives from arbitrarily large unlabeled pools. Our internal accounting for the roughly 10k-step runs across the three backbones is approximately 3.2k GPU-hours total, while deployment adds only the learned LoRA adapters.

\section{Additional Diagnostic Analyses of Core Mechanisms}
\label{sec:sup_analysis}

These additional diagnostic plots support two central design claims of the framework. First, they show why Solver Token Entropy (STE) is useful in addition to sample-level self-consistency by remaining informative when answer agreement alone becomes ambiguous. Second, they show how the shared \emph{Solver} links visual understanding and image generation by providing internal generation-side evaluation signals that strengthen as understanding improves over training.

\begin{figure*}[!htb]
  \centering
  \includegraphics[width=\linewidth]{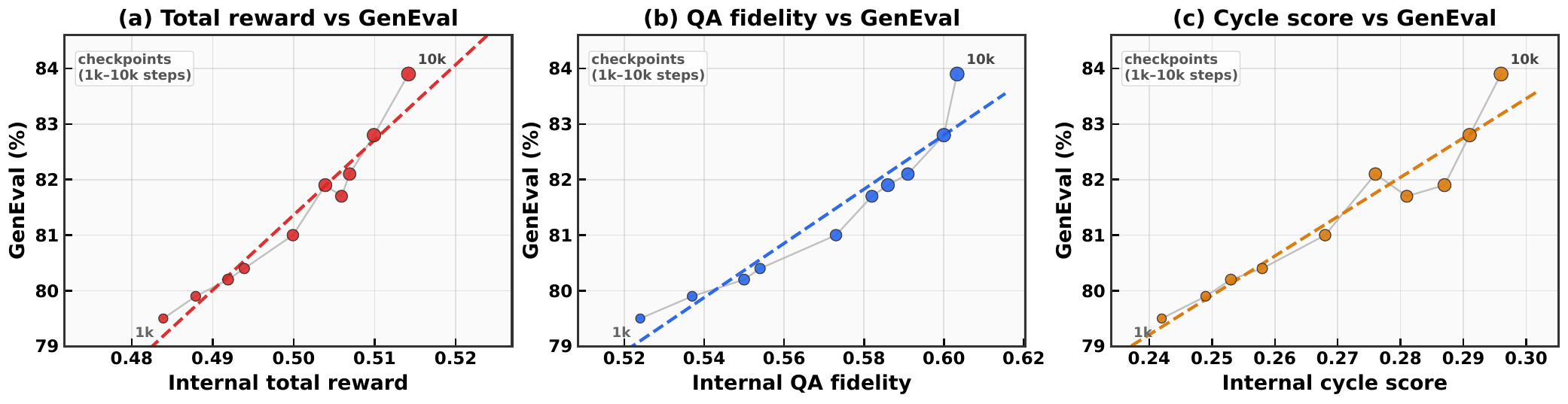}
  \caption{Checkpoint-level relationship between internal generation-side scores and GenEval overall~\cite{ghosh2024geneval}. Panels compare total reward, QA fidelity, and cycle score against GenEval across checkpoints spanning the roughly 10k-step training horizon. All three views follow a positive trend, with QA fidelity showing the clearest alignment. This supports the claim that the shared \emph{Solver} provides a meaningful internal evaluation signal for image generation, while cycle consistency remains complementary but noisier.}
  \label{fig:sup_reward_calibration}
\end{figure*}

Figure~\ref{fig:sup_reward_calibration} examines whether the internal generation-side scores used during training are aligned with external generation quality. The clearest relationship appears for QA fidelity, while total reward and cycle consistency follow the same direction with more variation. This supports the intended role of the shared \emph{Solver} as an internal evaluator whose signals remain informative at the checkpoint level rather than acting as an arbitrary self-reinforcing reward.

\begin{figure*}[!htb]
  \centering
  \includegraphics[width=\linewidth]{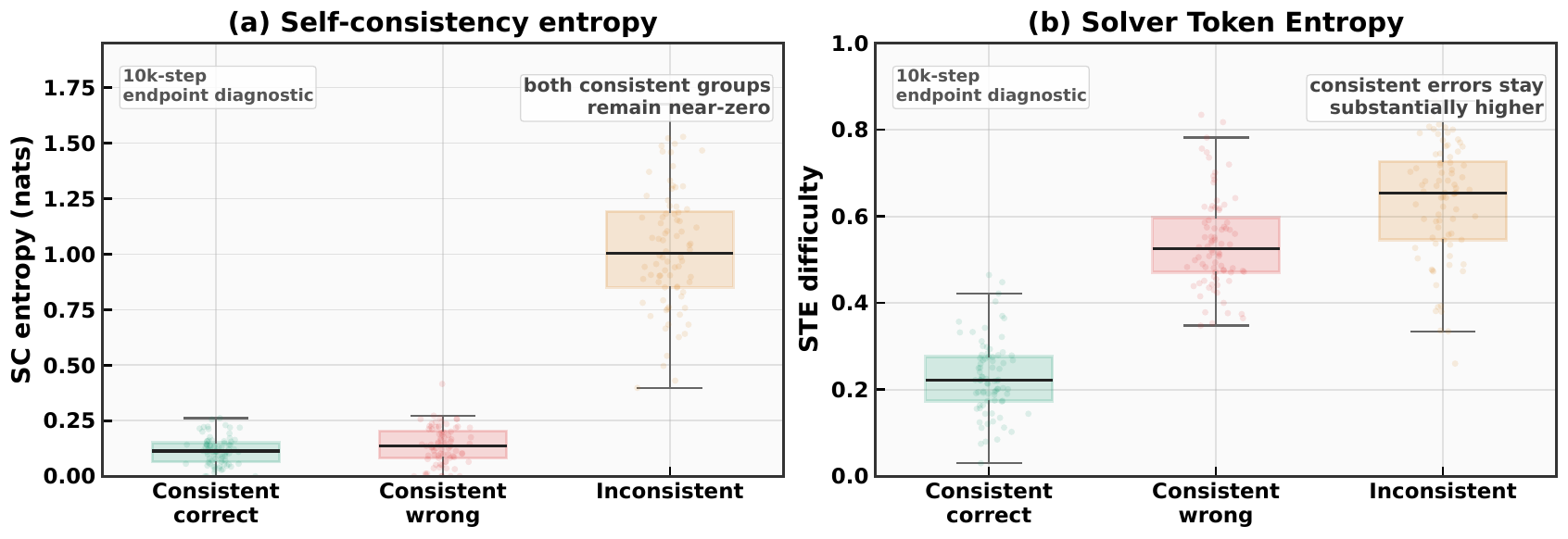}
  \caption{Diagnostic comparison of self-consistency entropy and Solver Token Entropy across three answer-behavior groups. Self-consistency entropy stays near zero for both consistent-correct and consistent-wrong cases, whereas STE remains substantially higher on consistent-wrong cases. This supports the claim that token-level uncertainty helps resolve the degenerate-signal regime in which sample-level agreement alone cannot distinguish confident correctness from confident error.}
  \label{fig:sup_ste_blindspot}
\end{figure*}

Figure~\ref{fig:sup_ste_blindspot} isolates the failure mode that motivates STE. Sample-level self-consistency entropy remains close to zero not only when the model is consistently correct, but also when it is consistently wrong. STE separates these two cases more clearly, showing why token-level uncertainty is needed to avoid collapsed agreement signals during frontier-seeking question selection.

\begin{figure*}[!htb]
  \centering
  \includegraphics[width=\linewidth]{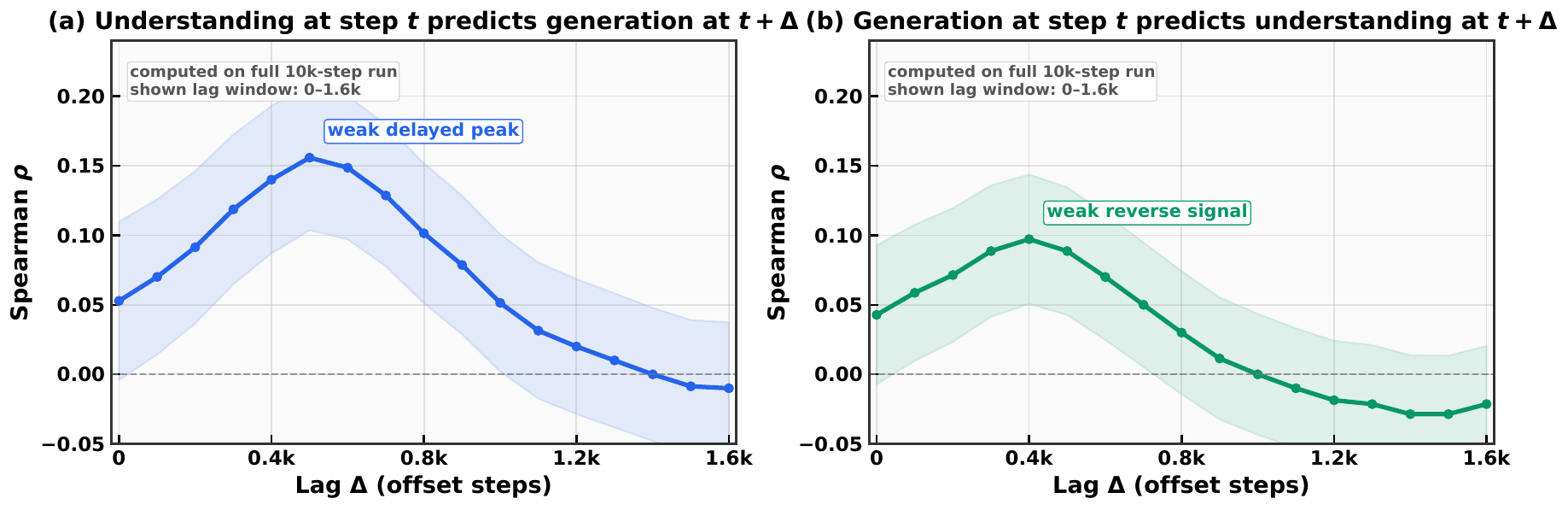}
\caption{Lagged correlation between understanding-side and generation-side signals over training. The left panel shows understanding-side metrics at step $t$ versus generation-side metrics at step $t+\Delta$, and the right panel shows the reverse direction. The understanding-to-generation direction exhibits the stronger peak, consistent with asymmetric coupling through the shared \emph{Solver}. The reverse-direction correlations are weaker and should be read as shared-schedule correlations, not as evidence of direct generation-to-understanding gradient feedback.}
  \label{fig:sup_lagged_coupling}
\end{figure*}

Figure~\ref{fig:sup_lagged_coupling} provides a complementary view of cross-loop interaction. The stronger understanding-to-generation peak suggests that improvements in visual understanding precede stronger generation-side evaluation signals, which is consistent with the design in which the shared \emph{Solver} serves as both learner and evaluator. We treat this figure as supporting evidence for the coupling claim alongside the main loop-ablation results rather than as a stand-alone causal test.

\section{Prompt Templates}
\label{sec:sup_prompts}

Table~\ref{tab:sup_prompt_families} summarizes the five prompt families used in BLIP3o self-evolving training~\cite{chen2025blip3o}. BAGEL~\cite{deng2025bagel} and VARGPT-v1.1~\cite{zhuang2025vargpt} use the same role decomposition and reward logic, but follow each backbone's native chat wrapper and generation API, so only the surface prompt wrapper differs while the underlying instructions remain analogous. All multimodal prompts are wrapped in a standard chat template with an image placeholder followed by the text prompt; text-only prompts omit the image token.

\begin{table}[!htb]
  \centering
  \small
  \setlength{\tabcolsep}{4pt}
  \renewcommand{\arraystretch}{1.12}
  \caption{Prompt families used in BLIP3o self-evolving training. The table organizes prompts by role and training phase, separating question proposal, answer generation, captioning, image generation, and generation-spec construction.}
  \label{tab:sup_prompt_families}
  \begin{tabular}{@{}p{0.13\linewidth}p{0.22\linewidth}p{0.57\linewidth}@{}}
    \toprule
    Role & Invocation & Prompt content \\
    \midrule
    Proposer & Visual understanding & Fixed adversarial question-writing preamble enforcing multi-domain reasoning, two-answer precision tests, and non-dominant grounding. Modulated at runtime by a sampled difficulty target (easy/medium/hard), a source-specific image hint, optional curriculum/anchor guidance, and fixed reasoning/task-card/strategy libraries. \\
    Solver & Visual understanding & One canonical answer prompt or one of seven PPS preambles (Table~\ref{tab:sup_pps_templates}), followed by shared rules enforcing 1--5 word answers and a runtime focus-hint line drawn from seven perceptual categories. \\
    Captioner & Image generation & Single-line prompt requesting a detailed image description. \\
    Generator & Image generation & Single-line wrapper converting a caption into a text-to-image instruction. \\
    Spec proposer & Image generation & Instruction block requesting one generation prompt and three compositional verification QA pairs in XML format, with an inserted difficulty target. \\
    \bottomrule
  \end{tabular}
\end{table}

\noindent\textbf{Proposer design.}
The proposer preamble enforces reasoning-first construction: each candidate question must use $\geq$2 reasoning domains (from a library of 7), include $\geq$1 non-relational domain, and target non-dominant visual details. A two-answer precision test is required to ensure the question has a single exact answer. The prompt concludes with an XML schema and a mandatory self-check that triggers rewrites for under-specified reasoning, vague alternatives, or questions that leak answer options. Runtime blocks modulate difficulty (three tiers), inject image-source hints (natural photo, text-in-scene, chart, relational scene), and optionally insert curriculum priorities or replay anchors.

\noindent\textbf{Exact proposer prompt anchor.}
BLIP3o uses a runtime multi-question proposer prompt. Its fixed prefix is:
\begin{promptbox}
\small
\ttfamily\raggedright
You are an Adversarial Fine-Grained Question Proposer.\par
GOAL: produce hard, visually grounded, objective questions using reasoning-first construction.\par
CRITICAL RULES:\par
- NEVER ask about the main/dominant subject, object, or largest text.\par
- Each question must use >=2 reasoning domains and include >=1 non-relational domain.\par
- Each question must include a valid two-answer precision test with distinct concrete alternatives.\par
- Final question must have one exact answer grounded in visible evidence.\par
- Use DISTINCT strategy codes across questions.\par
...\par
Output XML only:\par
<questions> ... </questions>
\end{promptbox}
The trainer then appends the exact difficulty block, dataset hint, optional curriculum/replay anchors, reasoning-domain block, task-card library, and strategy library from the prompt builder.

\noindent\textbf{Solver and prompt-perturbed sampling (PPS).}
Table~\ref{tab:sup_pps_templates} lists the seven PPS preambles. All variants share the same answer-format rules (1--5 word answers, no vague terms, XML output) and a runtime focus-hint drawn from a pool of seven perceptual categories. This design ensures answer diversity comes from framing rather than rule changes.

\begin{table}[!htb]
  \centering
  \small
  \setlength{\tabcolsep}{4pt}
  \renewcommand{\arraystretch}{1.05}
  \caption{Prompt-perturbed sampling (PPS) variants used by the Solver. Only the preamble changes across variants; the answer-format rules and the question remain fixed so that diversity comes from framing rather than from altered constraints.}
  \label{tab:sup_pps_templates}
  \begin{tabular}{@{}cp{0.82\linewidth}@{}}
    \toprule
    ID & PPS preamble \\
    \midrule
    1 & \texttt{You are a precise vision-language solver. Answer the question using only the provided image.} \\
    2 & \texttt{Look at the image carefully and provide a precise answer. Base your response solely on what is visible.} \\
    3 & \texttt{You are a visual analyst examining this image. Provide a factual answer derived from the visual evidence.} \\
    4 & \texttt{Study the image and answer the following question directly. Use only observable evidence from the image.} \\
    5 & \texttt{As an image examiner, answer the question below. Your answer must be grounded in what the image shows.} \\
    6 & \texttt{Based on the image provided, give a brief factual answer. Respond with only what you can verify visually.} \\
    7 & \texttt{Examine the visual evidence in this image. Answer the question using only observable details.} \\
    \bottomrule
  \end{tabular}
\end{table}

\noindent\textbf{Exact canonical solver prompt.}
The baseline solver call uses the following exact template, where \texttt{<focus hint>} and \texttt{<question text>} are filled at runtime:
\begin{promptbox}
\small
\ttfamily\raggedright
You are a precise vision-language solver.\par
Answer the question using only the provided image.\par
Rules:\par
- Your answer MUST be 1-5 words only. No full sentences.\par
- Give only the core answer, not an explanation.\par
- The answer must be concrete and exact, not vague.\par
- Focus mode for this sample: <focus hint>.\par
  Prefer evidence consistent with this focus.\par
- If the question asks `how many' or `number of', answer with a single integer.\par
- Never output vague count words or uncertainty phrases.\par
- Return only the final answer inside XML: <answer>...</answer>\par
Question: <question text>
\end{promptbox}
The seven runtime focus hints cover global scene layout, fine text and symbols, occlusion boundaries, left-right spatial relations, counting visible instances, color and texture evidence, and object interaction cues.

\noindent\textbf{Generation prompts.}
The captioner uses the exact prompt \texttt{Describe this image in detail.} The generator wrapper is exactly \texttt{Please generate image based on the following caption: <caption>}. The spec proposer asks for one image-grounded generation prompt and three compositional verification QA pairs (each requiring $\geq$2 visual cues) in XML format. Rejected specs are retried with the same schema plus the rejection reason.

\noindent\textbf{Compact exact generation-spec schema.}
The generation-spec prompt requires XML only in the following fixed structure:
\begin{promptbox}
\small
\texttt{<prompt>...</prompt>}\\
\texttt{<spec>}\\
\texttt{\hspace*{1em}<qa><question>...</question><expected>...</expected></qa>}\\
\texttt{\hspace*{1em}<qa><question>...</question><expected>...</expected></qa>}\\
\texttt{\hspace*{1em}<qa><question>...</question><expected>...</expected></qa>}\\
\texttt{</spec>}
\end{promptbox}
The retry prompt keeps the same XML schema and prepends three lines:
\begin{promptbox}
\small
\texttt{Your previous spec was rejected. Produce a better one.}\\
\texttt{Previous prompt:}\\
\texttt{Rejection reason:}
\end{promptbox}

\section{Additional Qualitative Results}
\label{sec:sup_qualitative}

\begin{figure*}[!htb]
  \centering
  \includegraphics[width=\linewidth,height=0.82\textheight,keepaspectratio]{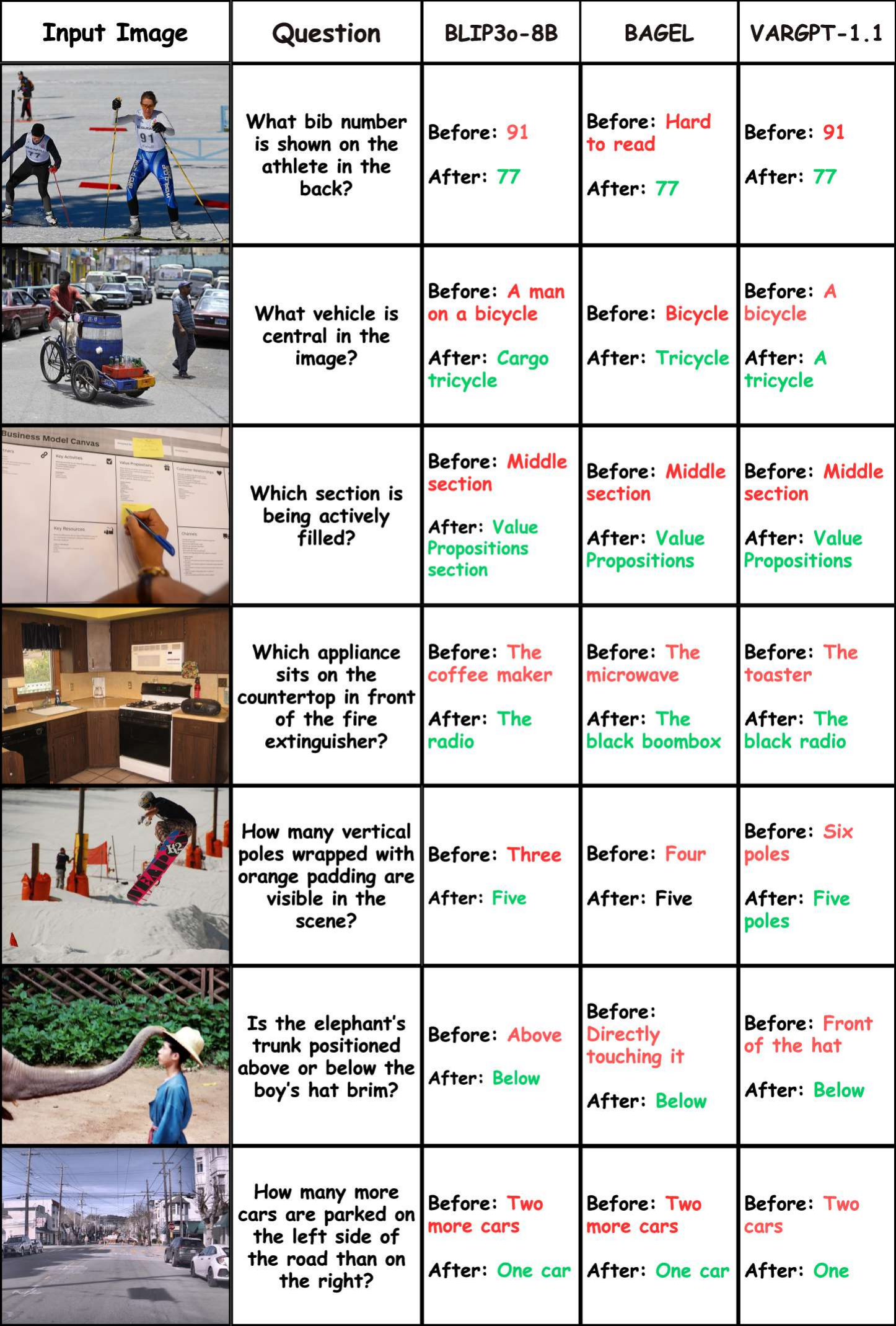}
  \caption{Visual understanding qualitative comparison. Baseline versus self-evolved answers for seven text-grounded, counting, and local-relation questions across the three backbones. Red marks baseline errors and green marks corrected post-training answers.}
  \label{fig:sup_understanding}
\end{figure*}

\begin{figure*}[!htb]
  \centering
  \includegraphics[width=\linewidth,height=0.82\textheight,keepaspectratio]{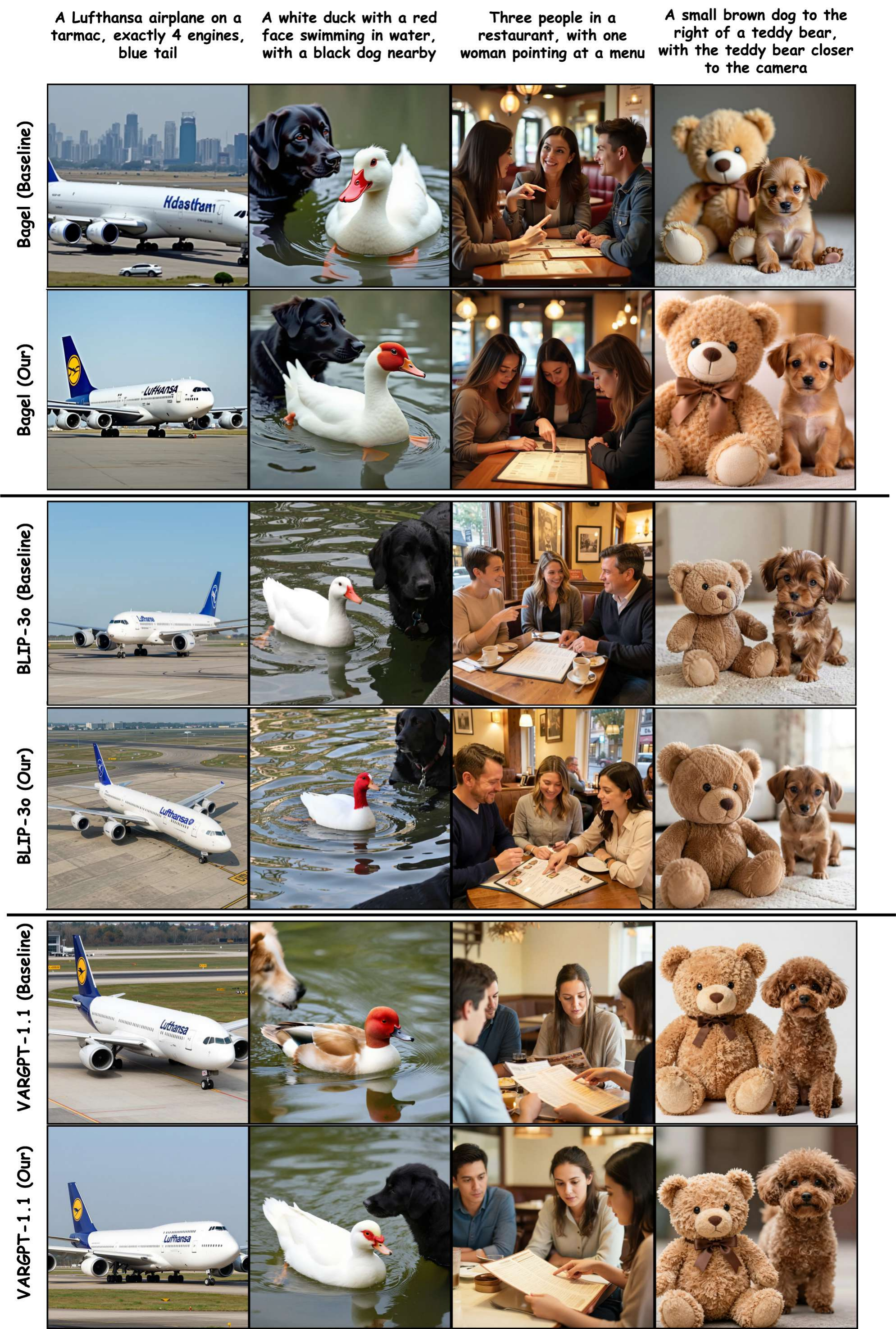}
  \caption{Natural-scene image generation qualitative comparison. Baseline versus self-evolved generations for four compositional prompts across the three backbones. The cases probe airline/logo-color binding, local color grounding in a two-entity scene, multi-person interaction around a menu, and relative positioning between two objects.}
  \label{fig:sup_generation}
\end{figure*}

\begin{figure*}[!htb]
  \centering
  \includegraphics[width=\linewidth,height=0.82\textheight,keepaspectratio]{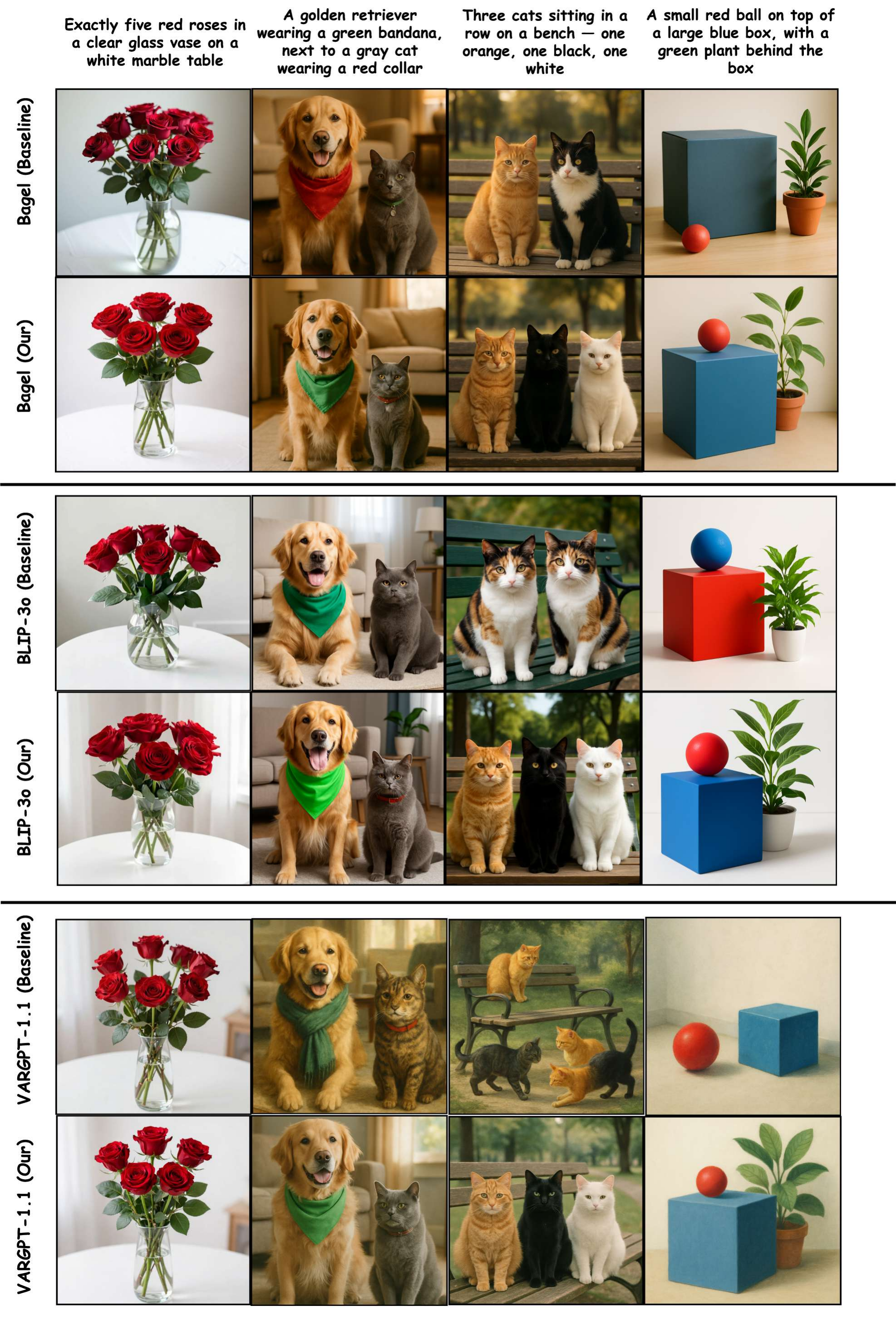}
  \caption{Controlled GenEval-style image generation qualitative comparison. Baseline versus self-evolved generations for four prompts spanning exact counting, cross-object attribute binding, multi-object composition, and spatial arrangement across the three backbones. These cases make improvements in count, attribute assignment, and relative placement visually verifiable.}
  \label{fig:sup_generation_2}
\end{figure*}

We illustrate before/after improvements across the three backbones for both understanding and generation. Figure~\ref{fig:sup_understanding} shows seven understanding cases spanning text grounding, counting, spatial reasoning, and local object identification. Figure~\ref{fig:sup_generation} presents four natural compositional generation prompts involving airline/logo-color binding, local color grounding in a two-entity scene, multi-person interaction, and relative positioning. Figure~\ref{fig:sup_generation_2} complements this with four controlled GenEval-style prompts that isolate counting, cross-object attribute binding, multi-object composition, and spatial arrangement.


\clearpage

\end{document}